\documentclass[11pt]{article}
\usepackage[english]{babel}
\usepackage{csquotes}
\usepackage[a4paper,top=2.54cm,bottom=2.54cm,left=2.54cm,right=2.54cm]{geometry}
\usepackage{authblk}           
\usepackage[autocite=superscript,style=nature,]{biblatex} 
\usepackage{enumitem}          
\usepackage{pgfplots}          
\pgfplotsset{compat=1.18}
\usepackage{multicol}          
\makeatletter
\let\@parboxrestore\relax
\makeatother
\usepackage[export]{adjustbox} 
\usepackage{wrapfig}           
\usepackage{capt-of}
\usepackage{caption}
\usepackage[version=4]{mhchem}
\usepackage{booktabs}          
\usepackage{threeparttable}
\usepackage{graphicx}          
\usepackage{amsmath}           
\usepackage{esvect}            
\usepackage{amssymb}           
\usepackage{hyperref}          

\usepackage{xcolor}
\usepackage{listings}
\usepackage{float}

\newfloat{listing}{htbp}{lol}
\floatname{listing}{Listing}

\lstdefinelanguage{yaml}{
  sensitive=false,
  comment=[l]{\#},
  commentstyle=\color{gray}\ttfamily,
  morestring=[b]",
  morestring=[b]',
  stringstyle=\color{purple},
  keywords={true,false,null,yes,no},
  keywordstyle=\color{blue}\bfseries
}

\title{\vspace{-2em}GEqTrain: A Configuration-Driven Framework for Retargeting Equivariant Graph Neural Networks Across 3D Scientific Tasks}

\author[1]{Daniele Angioletti}
\author[1]{Marco Nobile}
\author[1]{Vittorio Limongelli\thanks{Correspondence to: \href{mailto:vittorio.limongelli@usi.ch}{vittoriolimongelli@gmail.com}}}
\affil[1]{Faculty of Biomedical Sciences, Euler Institute, Universit\'a della Svizzera italiana, 6900 Lugano, Switzerland}

\date{}

\begin{document}

\maketitle

\setcounter{footnote}{0} 

\begin{abstract}

\noindent Equivariant graph neural networks provide a powerful modeling language for three-dimensional scientific data, but their reuse is often limited by implementations tied to specific tasks, outputs, and training regimes.
We present \textbf{GEqTrain}, a configuration-driven framework that separates dataset semantics, model composition, and training objectives.
Raw data are mapped to typed node-, edge-, and graph-level fields, while model stacks,
losses, and training workflows are assembled declaratively through Hydra configurations.
A shared equivariant backbone and training infrastructure can therefore be \textbf{retargeted to a new task primarily through configuration}.
We demonstrate this flexibility on three different problems handled within one software stack: coarse-grained-to-atomistic backmapping of biomolecular systems, prediction of NMR chemical shifts in molecular solids, and equivariant generative modeling.
Our aim is not to surpass individually optimized task-specific systems, but to show that a \textbf{shared representation and training infrastructure can achieve competitive accuracy across qualitatively different tasks at the cost of a configuration change}.
We further introduce \textbf{GEqDiff}, a generative extension based on equivariant flow matching. GEqDiff treats user-defined equivariant fields as first-class generation targets, jointly transporting Cartesian positions and non-scalar node fields spanning representations up to \(\ell=3\) within a single equivariant flow. We validate this capability on a controlled synthetic benchmark inspired by protein secondary-structure motifs, showing that fields with heterogeneous transformation properties can be reconstructed jointly and with high fidelity.
By reducing the software overhead of moving between predictive and generative, scalar and tensorial settings, GEqTrain aims to make equivariant modeling more reproducible, extensible, and reusable.
\end{abstract}

\section{Introduction}

Learning from three-dimensional scientific data is central to modern computational chemistry, materials science, and biomolecular modeling. In many of these settings, the targets of interest depend not only on composition and connectivity, but also on local geometry, conformation, and interactions in real space. This is evident in tasks such as atomistic property prediction, experimental-observable regression, coarse-grained backmapping, and structure generation, where the relevant quantities are constrained by Euclidean symmetries and by the physical organization of matter in three dimensions \cite{Haghighatlari2022, Stokes2020, Gaudelet2021, Abate2023, Unke2021, Batzner2022, Musaelian2023, Ko2021, Reiser2022, Xie2018, Jumper2021, Bronstein2021}.

Equivariant graph neural networks provide a natural modeling language for this regime. By construction, they encode how features should transform under rotations, translations, and, when appropriate, inversion, so that geometric information is processed as physical structure rather than as an arbitrary coordinate choice. This principle has enabled substantial progress across molecular property prediction, atomistic simulation, and protein modeling, and has produced a rapidly growing family of architectures with different trade-offs in expressivity, locality, and computational cost \cite{Satorras2021, schutt2017schnet, Gasteiger2022Directional, painn2021, Batzner2022, Musaelian2023, aykent2025gotennet, Baek2021}.

However, progress in model design has not fully translated into reusable research software. Many implementations remain tightly coupled to a specific task family, output type, or architectural template, making it difficult to change input semantics, move between scalar and equivariant targets, or reuse the same stack across supervised and generative workflows. The bottleneck is therefore not only architectural design, but software abstraction: equivariant learning frameworks must separate dataset semantics, geometric representations, model composition, and training objectives into modular, interchangeable components.

\textbf{GEqTrain} addresses this software gap as a configuration-driven framework for heterogeneous equivariant learning problems. Raw dataset fields are mapped to internal \textbf{typed node-, edge-, and graph-level fields}; geometry-derived quantities are constructed explicitly; and equivariant model stacks are assembled declaratively through Hydra-based configuration \cite{Yadan2019Hydra}. This design makes it possible to reuse the same framework across tasks that differ in input structure, target type, and training objective. Figure~\ref{fig:architecture_scheme} provides an overview of this configuration-driven workflow. GEqTrain emphasizes strictly local equivariant stacks, which are scientifically natural for many molecular interactions~\cite{kohn1996density,behler2007generalized} and enable memory-bounded chunked inference for node- and edge-local outputs, extending the accessible graph size under a fixed GPU-memory budget (Supplementary Fig. 1).

We emphasize that GEqTrain is a methodological contribution: the benchmarks in this work are used to test whether heterogeneous tasks can be expressed within one reproducible and configurable stack, not to argue that a modular framework should supersede task-specialized systems optimized for a single benchmark.

The present paper is biologically motivated, but intentionally broader in methodological scope than a single biomolecular application. Two case studies are drawn directly from molecular science: hierarchical coarse-grained backmapping of biomolecular systems and prediction of NMR chemical shifts in molecular solids, including scalar and tensorial local observables. To probe extensibility toward generative modeling settings relevant to ligand-oriented 3D design, we further introduce a controlled synthetic benchmark in which node states include Cartesian coordinates together with scalar and higher-order equivariant shape descriptors and dipole-like vectors. Although synthetic, this task isolates geometric ingredients that recur in molecular generation, namely the joint treatment of placement, orientation, local shape, and auxiliary descriptors within a single equivariant learning problem.

Our contributions are as follows:
\begin{itemize}
    \item We present GEqTrain, a configuration-driven framework for equivariant graph learning that separates dataset semantics, geometric representation, model composition, and training objectives within a unified software stack.
    \item We introduce a modular local architecture built from reusable invariant and equivariant primitives, supporting scalar, equivariant, and mixed-output prediction settings while remaining compatible with scalable execution strategies.
    \item We show that the same framework-level abstractions can support both predictive and generative workflows, rather than requiring separate software ecosystems for supervised regression and 3D generation.
    \item We demonstrate this versatility on three substantially different settings: hierarchical biomolecular backmapping, NMR chemical-shift prediction with scalar and tensorial targets, and a controlled equivariant generative benchmark potentially useful for 3D structure-based ligand design.
\end{itemize}

\section{Related Work}

Equivariant and geometry-aware graph neural networks have become central to modern molecular machine learning, spanning small-molecule property prediction, tensorial observables, and atomistic energy-and-force modeling. Early 3D graph models established the value of local geometric representations, while later architectures increasingly moved from invariant message passing toward explicitly equivariant treatments of vectorial and higher-order information. Representative milestones in this progression include SchNet, DimeNet, PaiNN, NequIP, Allegro, and related higher-order equivariant models, which together illustrate the shift from distance-based graph learning to tensorial and strictly local equivariant representations \cite{Gilmer2017, schutt2017schnet, Gasteiger2022Directional, painn2021, Batzner2022, Musaelian2023, Batatia2022MACE}.

From a software perspective, the field is also moving from isolated model implementations toward reusable and interoperable infrastructure. Existing efforts span several levels of abstraction. Libraries such as \texttt{e3nn} provide representation-theoretic primitives for equivariant neural networks, while atomistic-learning frameworks such as SchNetPack, TorchMD, and JAX MD provide reusable infrastructure for training, simulation, or differentiable molecular modeling. In parallel, the metatensor ecosystem developed by the COSMO community has explicitly targeted interoperability in atomistic machine learning: \texttt{metatensor} provides metadata-rich tensor containers for atomistic data, \texttt{metatomic} defines portable interfaces between machine-learning models and simulation engines, and \texttt{metatrain} provides a common training interface for atomistic models across architectures and targets~\cite{e3nn,Schutt2023SchNetPack2,Doerr2021TorchMD,Schoenholz2020JAXMD,Bigi2026Metatensor,MetatrainSoftware}. Highly optimized model-specific stacks such as NequIP, Allegro, and MACE have further demonstrated how software design can make equivariant interatomic potentials practical at scale~\cite{Batzner2022,Musaelian2023,Batatia2022MACE}.
GEqTrain is complementary to these efforts. Rather than introducing a new low-level equivariant algebra or a single task-specialized architecture, it focuses on the interface between typed scientific data semantics, modular equivariant computation, and task-level objectives, with the aim of expressing heterogeneous predictive and generative workflows within one configuration-driven framework.

A particularly relevant application area for our first case study is coarse-grained backmapping. Classical backmapping pipelines often rely on geometric reconstruction rules followed by relaxation, and remain valuable in many workflows, but recent machine-learning approaches increasingly formulate backmapping as a conditional generative problem. FlowBack uses flow matching to map coarse-grained configurations to atomistic structure distributions, CGBack employs diffusion-based reconstruction for large and complex coarse-grained biomolecular systems, and MSBack explores constrained diffusion for highly coarse-grained proteins \cite{Jones2025Flowback, Berlaga2025FlowbackAdjoint, UgarteLaTorre2025, Waltmann2025}. These methods highlight a broader shift toward generative multiscale reconstruction in settings where stereochemistry, packing, and conformational diversity must be recovered directly from coarse variables. In this context, our backmapping study is used to test whether GEqTrain can express a complex task-specific reconstruction pipeline while preserving a reusable equivariant feature-extraction stack.

NMR chemical-shift prediction provides a different test case: instead of reconstructing coordinates, the model must learn a local structure--observable map whose target is physically defined by the electronic response of a nucleus to an external magnetic field. Earlier graph-based and 3D message-passing approaches already showed that geometric representations can recover local effects relevant to chemical shifts \cite{Yang2021Predicting, Han2024}. More recent models have improved performance through larger datasets, task-specific architectures, ensemble strategies, and pretraining, including GT-NMR, GeqShift, and ensemble message-passing approaches \cite{Chen2024, Bankestad2024, Williamson2024}. In molecular solids, the ShiftML line of work provides a particularly relevant benchmark family: the original CSD-2k/CSD-500 setting evaluates isotropic shielding prediction under controlled DFT-relaxed crystal conditions, whereas ShiftML3 expands the scope to a larger dataset and to full shielding tensors using a specialized PET/nanoPET ensemble \cite{Kellner2025}.

The generative case study is motivated by a different literature at the interface of biomolecular interaction modeling and ligand-oriented 3D design. Recent work has made rapid progress in deep generative and interaction-aware models for protein--ligand structure prediction and design, including diffusion-based docking and structure-generation systems as well as broader multimodal structure predictors \cite{Lu2024, Qiao2024, Bryant2024, Abramson2024, Schneuing2024, Passaro2025}.
Recent generative models have begun to exploit higher-order equivariant representations more directly.
Symphony, for example, combines \(E(3)\)-equivariant message passing with spherical-harmonic projections to parameterize autoregressive distributions over atomic placements
\cite{Daigavane2024Symphony}.
Clifford Diffusion Models instead encode molecular structures as graded Clifford multivectors and diffuse a joint latent state containing vector and higher-grade components \cite{Liu2025CliffordDiffusion}. In both cases, the additional equivariant representations primarily support the generation of molecular coordinates.
The GEqDiff benchmark examines a complementary setting in which equivariant fields are not only internal representations, but explicit generative targets with prescribed irreps and field-specific velocity predictions.
We introduce a synthetic LEGO benchmark that isolates geometric ingredients relevant to three-dimensional molecular modeling: Cartesian positions, local spherical-harmonic shape descriptors, and vector-valued directional attributes. These fields are jointly transported within a single flow-matching objective while remaining separately declared, supervised, and evaluated. The benchmark therefore tests whether the typed abstractions of GEqTrain extend consistently from predictive learning to mixed-field generation, providing a controlled methodological basis for future applications involving molecular docking and design.

\section{Model Configuration and Architecture}
\label{sec:architecture}

\begin{figure}[htbp]
    \centering
    \includegraphics[width=1.0\textwidth]{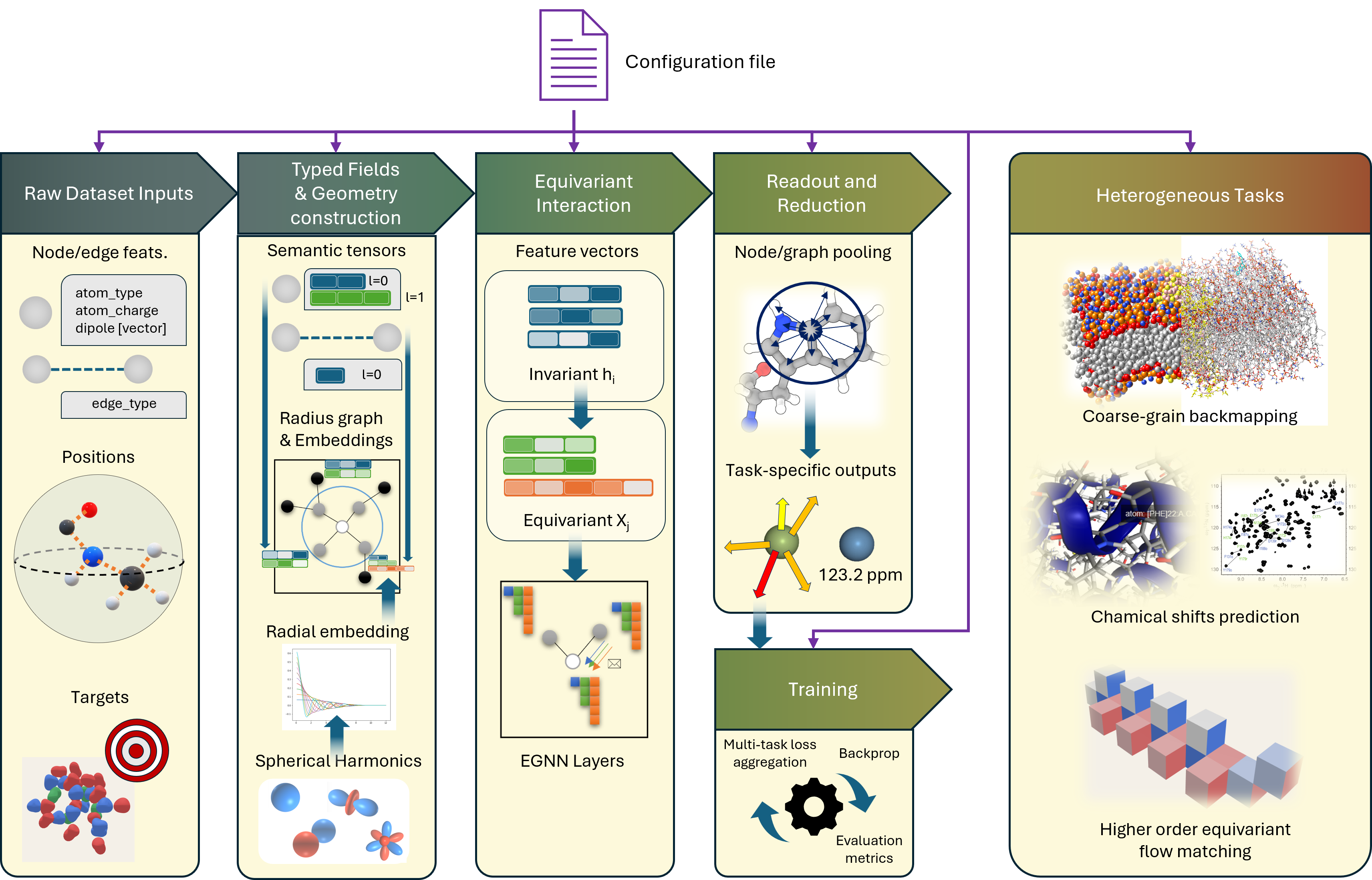}
    \caption{Overview of the GEqTrain workflow. Raw dataset tensors are mapped to typed node-, edge-, and graph-level fields through configuration. Geometry-derived attributes, including radial and angular edge embeddings, are then constructed on a radius graph and
    processed by a reusable local equivariant interaction backbone. The resulting latent features are reduced and decoded into task-specific outputs, which feed into a unified training workflow handling multi-task loss aggregation, backpropagation, and metric evaluation to support heterogeneous applications such as backmapping, scalar property prediction, and equivariant generative modeling.}
    \label{fig:architecture_scheme}
\end{figure}

\textbf{GEqTrain} is a modular framework for building and training
equivariant graph neural networks on structured three-dimensional scientific
data. The empirical applications considered in this work are molecular and
biomolecular, while the software abstractions are defined independently of a
specific molecular target.
Its design is inspired by flexible frameworks such as NequIP and Allegro~\cite{Batzner2022, Musaelian2023}, while extending configurability beyond interatomic potentials to a broader range of prediction targets, data modalities, and training regimes.
At the core of GEqTrain is a Hydra-based configuration tree~\cite{Yadan2019Hydra} in which an experiment is assembled from dedicated configuration groups through Hydra's \texttt{defaults} mechanism.
This composition separates data semantics, model architecture, and optimization settings, while preserving a single reproducible experiment definition.
An overview of this organization is shown in Figure~\ref{fig:architecture_scheme}, which summarizes how GEqTrain maps raw scientific data to typed internal fields (i.e. node-, edge-, and graph-level fields), constructs geometric features, applies a local equivariant interaction stack, and exposes task-specific readouts within a single configurable workflow.

\subsection{Design principles and notation}
\label{subsec:design_notation}

The framework is built around two design principles.
First, all tensors exchanged between modules carry an explicit \emph{semantic role} (for example positions, node attributes, edge attributes, node features, or graph outputs).
Second, the transformation behavior of learned features is declared explicitly through irreducible representations of \(O(3)\), so that invariant and equivariant quantities can be processed within a unified interface.

\paragraph{Transformation conventions.}
Throughout this work, we use the \texttt{e3nn} \cite{e3nn} notation of irreducible representations (\emph{irreps}) of \(O(3)\).
An irrep is identified by an angular order \(\ell=0,1,2,\dots\) and a parity label \(e/o\), indicating whether the channel is even or odd under inversion.
The notation \texttt{mxle/o} denotes \(m\) copies of the corresponding irrep. 
In particular, \texttt{1x0e} denotes an invariant scalar channel, while \texttt{1x1o} and \texttt{1x2e} denote higher-order channels that transform equivariantly under rotations and inversion.

At the network level, GEqTrain is designed to respect \(E(3)\) symmetries of three-dimensional molecular systems. Translation dependence is removed by expressing geometry through relative coordinates, while rotational and inversion behavior is controlled through irreps-aware equivariant operations.
In the remainder of the manuscript, we use the term \emph{invariant} for \(\ell=0\) channels and \emph{equivariant} for non-scalar channels that transform according to higher-order irreps.

\paragraph{Notation.}
To keep the presentation uniform, we distinguish four levels of representation. Raw typed inputs are denoted by \(x_i\), \(x_{ij}\), and \(x_G\) for node-, edge-, and graph-level quantities.
Their embedded versions are denoted by \(a_i\), \(a_{ij}\), and \(a_G\). Geometry-derived edge attributes are written separately as an invariant radial embedding \(\rho_{ij}\) and an equivariant angular embedding \(\mathbf y_{ij}^{\mathrm{sh}}\).
Learned latent features inside the interaction stack are written as invariant edge states \(h_{ij}^{(\ell)}\) and equivariant edge states \(\mathbf z_{ij}^{(\ell)}\) at layer \(\ell\).

\subsection{Config-driven experiment composition}
\label{subsec:config_composition}

A GEqTrain experiment is assembled from Hydra configuration groups specifying the dataset interface, the ordered model stack, and the optimization setup. The top-level experiment file therefore remains intentionally compact: it declares which data definition, model definition, and training recipe are combined.
This organization makes the experiment definition reproducible while avoiding hard-coded coupling between dataset conventions and model implementation.

\subsection{Typed molecular representation}
\label{subsec:typed_representation}

Given a molecular structure with atomic positions \(\mathbf r_i \in \mathbb R^3\), GEqTrain constructs a directed radius graph
\[
\mathcal G=(\mathcal V,\mathcal E), \qquad
(i,j)\in\mathcal E \iff i\neq j \ \text{and}\ \|\mathbf r_j-\mathbf r_i\|\le r_{\max}.
\]
Using directed edges is convenient because many modules are edge-centric and distinguish source and target roles explicitly.

The framework does not assume a unique dataset convention. Instead, raw tensors from a file or dataloader are mapped to internal typed fields through configuration.
For example, positions may be mapped to \texttt{pos}, categorical atomic identities to \texttt{node\_types}, graph-level targets to fields such as \texttt{energy}, and node-level targets to fields such as \texttt{forces} or \texttt{chemical\_shifts}.
This typed-field abstraction is central to the framework: the model stack consumes named semantic fields rather than dataset-specific tensor layouts.

\begin{listing}[htbp]
\caption{Illustrative GEqTrain dataset mapping for a scalar node-level task.}
\label{lst:data_compact}
\begin{lstlisting}[language=yaml,gobble=4]
    train_dataset_list:
      - dataset: npz
        dataset_input: /path/to/train.npz
        key_mapping:
          coords: pos
          atom_types: node_types
          chemical_shifts: cs
          Lattice: cell

    node_fields:
      - cs

    fixed_fields:
      - node_types
\end{lstlisting}
\end{listing}

The data configuration in Listing~\ref{lst:data_compact} defines the interface between raw files and the internal typed fields used by GEqTrain. Through \texttt{key\_mapping}, raw dataset keys are mapped to framework-level semantic names such as \texttt{pos}, \texttt{node\_types}, or task-specific targets. The distinction between \texttt{node\_fields}, \texttt{graph\_fields}, and \texttt{fixed\_fields} specifies whether a quantity is interpreted as frame-dependent, graph-level, or constant for all frames in the corresponding data source.

Datasets may be organized as collections of NPZ files, for example one system per file with one or more frames, or packed into a single NPZ archive using padded arrays together with companion \texttt{\_\_mask\_\_} fields to indicate valid entries. We defer the detailed discussion of storage choices to the Supplementary Information.

\paragraph{Geometric edge features.}
For each directed edge \((i,j)\), the relative displacement, distance, and unit direction are
\[
\mathbf r_{ij}=\mathbf r_j-\mathbf r_i,\qquad
d_{ij}=\|\mathbf r_{ij}\|,\qquad
\hat{\mathbf r}_{ij}=\frac{\mathbf r_{ij}}{d_{ij}}.
\]
GEqTrain represents local geometry through two complementary edge attributes. The first is an invariant radial embedding
\[
\rho_{ij}=\phi(d_{ij}),
\]
where \(\phi\) denotes a learnable or fixed radial basis expansion of interatomic distance. The second is an equivariant angular embedding
\[
\mathbf y_{ij}^{\mathrm{sh}}
=
\bigoplus_{\ell \le \ell_{\max}} Y^{(l)}(\hat{\mathbf r}_{ij}),
\]
obtained from spherical harmonics evaluated on the edge direction. The radial term provides scalar metric information, whereas the spherical harmonics provide angular information in irreps-compatible form.


\paragraph{Typed input attributes.}
After raw tensors have been mapped to semantic field names, GEqTrain uses a second configuration layer to declare how these fields enter the equivariant model.
This declaration separates two independent aspects of an input: the \emph{domain} on which it is defined, such as node, edge, or graph, and its \emph{transformation type}, specified through scalar embeddings or explicit irreps.

Invariant categorical or numerical node fields are declared under \texttt{node\_attributes} and are embedded into scalar channels.
Examples include atom types, sequence indices, masks, conditioning vectors, or other quantities that should not rotate with the molecular frame.
Tensor-valued node fields are declared separately under \texttt{eq\_node\_attributes}; these fields must specify their irreducible representations and are embedded in a way that preserves their equivariant transformation behavior.
Thus, \texttt{node\_attributes} and \texttt{eq\_node\_attributes} may both describe node-level quantities, but they differ in how those quantities transform and are processed by the equivariant stack.

Listing~\ref{lst:attribute_declaration} shows this distinction for the LEGO flow-matching benchmark.

\begin{listing}[htbp]
\caption{
Declaration of invariant and equivariant node attributes in the LEGO flow-matching benchmark.
Both blocks define node-level inputs, but \texttt{node\_attributes} are embedded as invariant scalar information, whereas \texttt{eq\_node\_attributes} declares tensor-valued inputs with explicit irreps.
}
\label{lst:attribute_declaration}
\begin{lstlisting}[language=yaml,gobble=4]
    node_attributes:
      sequence_position:
        attribute_type: categorical
        embedding_mode: positional
        num_types: 48
        embedding_dimensionality: 16
      branch_kind:
        attribute_type: categorical
        embedding_mode: one_hot
        num_types: 4
      ligand_mask:
        attribute_type: numerical
        embedding_dimensionality: 1
      pocket_mask:
        attribute_type: numerical
        embedding_dimensionality: 1
      conditioning:
        attribute_type: numerical
        embedding_dimensionality: 64

    eq_node_attributes:
      shape_features:
        attribute_type: numerical
        irreps: 1x0e + 1x1o + 1x2e + 1x3o
        embedding_dimensionality: 16
      dipole_direction:
        attribute_type: numerical
        irreps: 1x1o
        embedding_dimensionality: 3
\end{lstlisting}
\end{listing}

\subsection{Reference strictly local architecture}
\label{subsec:reference_local_architecture}

The default stack used to illustrate GEqTrain in this work follows a strictly local, edge-centric design. It can be summarized in four stages: input preparation, equivariant interaction on the cutoff graph, feature reduction, and task-specific readout.
This is a \emph{reference} architecture rather than the only architecture supported by the framework.

\paragraph{Input preparation.}
The first stage prepares all fields required by the interaction backbone. Gradient tracking can be enabled for differentiable inputs such as positions when derivative targets will later be recovered from a scalar prediction. User-defined node, edge, and graph attributes are embedded into typed internal fields. In parallel, geometric edge features \(\rho_{ij}\) and \(\mathbf y_{ij}^{\mathrm{sh}}\) are constructed from the radius graph. After this stage, each edge carries the information needed for local equivariant processing: task-defined attributes together with invariant and angular geometric context.

\paragraph{Equivariant interaction on directed edges.}
The computational core is an \texttt{Interaction\allowbreak Module} composed of one or more \texttt{InteractionLayer}s. The reference backbone used in this work is strictly local: all updates are defined on the radius graph and depend only on edge-local attributes, geometric embeddings, and reductions over neighboring edges. This locality makes the interaction channels fixed by the cutoff graph, which keeps the model compatible with memory-controlled and chunked execution on large molecular systems, while leaving non-local or attention-based extensions as optional architectural choices.
At layer \(\ell\), the module maintains an invariant edge state \(h_{ij}^{(\ell)}\) and, when requested by the configuration, an equivariant edge state \(\mathbf z_{ij}^{(\ell)}\). At a high level, each interaction layer performs three operations:
\begin{align}
(h_{ij}^{(0)}, \mathbf z_{ij}^{(0)}) &= \Phi_{\mathrm{init}}(a_i, a_j, a_{ij}, \rho_{ij}, \mathbf y_{ij}^{\mathrm{sh}}), \\
\mathbf c_i^{(\ell)} &= \mathrm{Reduce}\Big(\{\,\Phi_{\mathrm{env}}(h_{ij}^{(\ell)}, \mathbf z_{ij}^{(\ell)}) : j \in \mathcal N(i)\,\}\Big), \\
(h_{ij}^{(\ell+1)}, \mathbf z_{ij}^{(\ell+1)}) &= \Phi_{\mathrm{int}}\!\left(h_{ij}^{(\ell)}, \mathbf z_{ij}^{(\ell)}, \mathbf c_i^{(\ell)}\right).
\end{align}
Here, \(\Phi_{\mathrm{init}}\) denotes the initialization of edge-local latent features from typed attributes and geometry, \(\Phi_{\mathrm{env}}\) builds edge contributions to a source-centered local environment, and \(\Phi_{\mathrm{int}}\) applies the actual equivariant update, typically involving irreps-aware tensor products and residual mixing. \(\mathcal N(i)\) represents the neighbourhood of atom \(i\).

\paragraph{Feature reduction.}
Because the backbone is edge-centric, downstream predictions usually require a change of semantic level. GEqTrain therefore provides explicit reduction stages. Edge features can be reduced to node-level representations, and node features can in turn be pooled to graph-level representations when the task requires a global output. This separation between interaction and reduction makes the stack easier to reuse across node-, edge-, and graph-level prediction problems.

\paragraph{Task-specific readout.}
Predictions are produced by readout modules that map a selected latent field to user-declared output irreps. This allows scalar, equivariant, or mixed outputs to be expressed within the same framework-level interface. For a graph property such as total energy, node-level contributions may first be predicted and then reduced to a graph scalar. For derivative observables such as forces, the framework can recover the target by differentiating a scalar prediction with respect to positions rather than by introducing a separate force-specific interaction backbone.

\section{Generative Extension via Equivariant Flow Matching}
\label{sec:generative_extension}

To demonstrate that GEqTrain extends beyond prediction, we introduce \textbf{GEqDiff}, a generative extension built on the same typed-tensor and modular-stack abstractions described above.
Although the framework can also express diffusion-style forward noising processes, here we focus on \textbf{equivariant flow matching}, which provides a natural continuous-time formulation for jointly transporting heterogeneous invariant and equivariant node fields. The novelty of GEqDiff is not the flow-matching formulation itself, which follows the standard linear-path construction, but the \textbf{object being generated}: GEqDiff transports \textit{higher-order equivariant node fields as first-class targets}, assigning each field (coordinates, spherical-harmonic shape descriptors, dipole-like vectors) its own velocity target and an irreps-matched readout head, so that heterogeneous invariant and equivariant quantities are denoised jointly and consistently. Diffusion-based alternatives are summarized in the Supplementary Information.

\subsection{Flow Matching over Mixed Equivariant Fields}
\label{subsec:flow_matching_mixed_fields}

A central design goal of GEqDiff is to support generative modeling over \emph{mixed} node states rather than Cartesian coordinates alone. Let
\[
\mathcal{X} = \{ \mathbf{x}^{(m)} \}_{m=1}^{M}
\]
denote the set of fields to be corrupted and generated jointly, where each field may correspond, for example, to Cartesian positions, invariant scalar descriptors, or equivariant tensor features. In the current implementation, these fields are specified explicitly through a list of \texttt{corrupt\_fields}, allowing each component to have its own velocity target and corruption settings. 

For each field \(\mathbf{x}^{(m)}\), GEqDiff defines a continuous interpolation between data and noise,
\[
\mathbf{x}^{(m)}_{\tau}
=
\alpha(\tau)\,\mathbf{x}^{(m)}
+
\sigma(\tau)\,\boldsymbol{\xi}^{(m)},
\qquad \tau \in [0,1],
\]
where \(\boldsymbol{\xi}^{(m)}\) is sampled noise of matching shape. The corresponding flow-matching target is the scheduler velocity
\[
\mathbf{u}^{(m)}_{\tau}
=
\dot{\alpha}(\tau)\,\mathbf{x}^{(m)}
+
\dot{\sigma}(\tau)\,\boldsymbol{\xi}^{(m)}.
\]
In the present implementation, the default \texttt{FlowMatchingScheduler} uses the linear path
\[
\alpha(\tau)=1-\tau,
\qquad
\sigma(\tau)=\tau,
\]
for which the target simplifies to
\[
\mathbf{u}^{(m)}_{\tau}
=
\boldsymbol{\xi}^{(m)}-\mathbf{x}^{(m)}.
\]
The training objective is then written as
\[
\mathcal{L}_{\mathrm{FM}}
=
\sum_{m=1}^{M}
\lambda_m\,
\mathbb{E}_{\tau,\mathbf{x},\boldsymbol{\xi}}
\left[
\left\|
\hat{\mathbf{u}}^{(m)}_{\tau}
-
\mathbf{u}^{(m)}_{\tau}
\right\|^2
\right].
\]
This formulation naturally accommodates mixed invariant and equivariant quantities while preserving the appropriate transformation behavior of each output channel. 

The field-wise corruption mechanism also supports masked generation. In particular, each field may be centered before corruption, selectively noised through a Boolean mask, and optionally left partially corrupted or uncorrupted outside the mask. 

\subsection{Integration into the GEqTrain Stack}
\label{subsec:flow_matching_stack_integration}

The transition from predictive to generative modeling is implemented by adding a generative front end to the GEqTrain stack. In the flow-matching setting, this front end samples a continuous time variable \(\tau\), constructs interpolated noisy states for selected fields, stores the corresponding velocity targets, and appends a sinusoidal time embedding to the conditioning features used by the downstream network. The equivariant interaction backbone, edge encodings, reductions, and readout modules are then reused without changing their basic interface.

This design separates three concerns that would otherwise be entangled in a task-specific implementation. First, the dataset and attribute interface declares which quantities exist and how they transform under rotations, e.g. positions, scalar node labels, vector fields, or higher-order tensorial descriptors. Second, the generative front end specifies which of these fields are transported during training, which entries are masked or kept fixed, and which velocity target is written for each transported field. Third, the readout heads predict field-specific velocities with output irreps matching the corresponding transported quantities. For example, coordinates and dipole directions require vector-valued \texttt{1x1o} velocity heads, whereas the LEGO shape descriptor requires a mixed scalar--tensorial head with irreps \texttt{1x0e + 1x1o + 1x2e + 1x3o}.

The practical consequence is that mixed-field generation is expressed as a configuration-level extension of the same typed-tensor stack used for predictive tasks. The model does not require a separate hard-coded architecture for each generated quantity: adding or removing a transported field amounts to changing the declared field, its noising/interpolation rule, and the corresponding irreps-aware readout. Compact YAML examples of the flow-matching front end and the matching velocity heads are reported in the Supplementary Information.

\section{Experiments and Results}

\subsection{Case Study: Hierarchical Coarse-Grained Backmapping with HEroBM}

To showcase GEqTrain's capabilities on a complex, multi-scale modeling challenge, we present its application in the HEroBM (Hierarchical Equivariant representation for optimised BackMapping) framework \cite{Angioletti2025}. HEroBM was originally developed and trained using GEqTrain, relying on its standard modules complemented by a custom module for the hierarchical reconstruction process. The results presented in this work, however, originated from an enhanced version of the HEroBM model. Specifically, the input features were improved from the original HEroBM publication to include sequence connectivity information (as detailed in the \hyperref[herobm:results]{Results paragraph}), demonstrating GEqTrain's flexibility in facilitating rapid model refinement and experimentation. 
\paragraph{Task:}
The task addressed by HEroBM is \textbf{coarse-grained (CG) backmapping}, which involves reconstructing full atomistic coordinates from reduced, coarse-grained representations of molecular systems. While CG models are invaluable for simulating large biomolecular systems over extended timescales, phenomena often inaccessible to full atomistic simulations due to computational cost, they inherently lose fine-grained structural details. Backmapping is therefore crucial for retrieving this atomistic information, allowing for detailed analysis of specific interactions (like hydrogen bonds), validation of CG simulation accuracy, and investigation of system properties that depend on atomic-level resolution.

HEroBM employs an SE(3)-equivariant graph neural network (EGNN) to predict the distance vectors of atoms relative to hierarchically defined anchor points (which can be CG beads or other atoms within the same bead).
This "one-shot" approach is designed to produce high-fidelity atomistic reconstructions directly from the CG model, with broad applicability across diverse molecular systems (proteins, lipids, small organic molecules) and various CG mapping schemes.
The motivation is to enable accurate and efficient backmapping for large, complex biochemical systems, which are often challenging for existing methods. 

\paragraph{Dataset:}
The HEroBM framework was benchmarked on a diverse range of molecular systems to showcase its broad applicability. For proteins, two main datasets were utilized:
\begin{itemize}
    \item The \textbf{PDB3k dataset}, derived from the Top8000 and PISCES sets \cite{Hintze2016Top8000,Wang2003Pisces}, consists of 2.9k protein structures filtered from the PDB29k dataset (prepared by the authors of cg2all \cite{Pang2024}). For this dataset, 2900 samples were used for training and 72 for validation, employing the Martini 3.0 coarse-grained mapping with a cutoff radius of 7.0 \AA{} and approximately 5 atoms per bead \cite{Souza2021}.
    \item The \textbf{PED (Protein Ensemble Database) dataset} \cite{Lazar2021PED} was also used, with training, validation, and testing splits identical to those in the GenZProt and DiAMoNDBack studies for consistency \cite{Yang2023,Jones2023}. This involved 3900 training samples and 80 validation samples, also using Martini 3.0 mapping, a 7.0 \AA{} cutoff, and an average of 5 atoms per bead. \end{itemize}
Beyond proteins, HEroBM's capabilities were extended to other molecular types:
\begin{itemize}
    \item \textbf{Lipid bilayers:} A dataset of lipid bilayers composed of 1-palmitoyl-2-oleoyl-sn-glycero-3-phosphocholine (POPC) and cholesterol (CHL) molecules was generated. This dataset comprised 100 frames from an atomistic simulation, with 10 frames (2920 lipid molecules) used for training and 5 frames (1460 lipid molecules) for validation. The Martini 3.0 mapping for lipids was applied, with a 10.0 \AA{} cutoff radius and an average of 6 atoms per bead. \item \textbf{Small organic molecules:} The A2A antagonist ligand ZMA was used as a test case. Atomistic MD trajectory frames, used to define the Martini CG mapping of ZMA, served as the structural input. The dataset included 1,000 entries, with 200 frames for training and 100 for validation, using a custom CG mapping, a 7.0 \AA{} cutoff, and an average of 4 atoms per bead. \end{itemize}
In all cases, the input to the model consisted of the coarse-grained representations (coordinates and bead types), and the target output was the reconstruction of the corresponding all-atom structures. 

\paragraph{Results:}\label{herobm:results}

\begin{table*}[!ht]
\begin{tabular}{@{}llcccc@{}}
\toprule
Dataset & & $CG2AT$ & $CG2ALL$ & $HEroBM$ & $HEroBM_{A2A}$ \\
\midrule
$PDB_{29k}$   & BB  & - & ${0.08\pm0.02}$ & $\mathbf{0.07\pm0.02}$ & $0.61\pm0.10$ \\
              & ALL & - & ${0.31\pm0.05}$ & $\mathbf{0.24\pm0.03}$ & $0.81\pm0.06$ \\
$PED_{00055}$ & BB & $0.88\pm0.05$ & $\mathbf{0.07\pm0.01}$ & $0.14\pm0.04$ & $0.64\pm0.07$ \\
              & SC & $1.36\pm0.03$ & $1.22\pm0.03$ & $\mathbf{0.54\pm0.05}$ & $1.00\pm0.03$ \\
$PED_{00090}$ & BB & $1.14\pm0.06$ & $\mathbf{0.09\pm0.01}$ & $0.15\pm0.05$ & $0.90\pm0.09$ \\
              & SC & $1.47\pm0.02$ & $1.27\pm0.02$ & $\mathbf{0.62\pm0.04}$ & $1.01\pm0.02$ \\
$PED_{00151}$ & BB & $0.93\pm0.05$ & $\mathbf{0.07\pm0.01}$ & $\mathbf{0.07\pm0.02}$ & $0.82\pm0.10$ \\
              & SC & $1.19\pm0.05$ & $1.06\pm0.03$ & $\mathbf{0.59\pm0.06}$ & $0.95\pm0.04$ \\
$PED_{00218}$ & BB & $0.81\pm0.02$ & $\mathbf{0.08\pm0.01}$ & $0.13\pm0.03$ & $0.62\pm0.09$ \\
              & SC & $1.30\pm0.03$ & $1.02\pm0.03$ & $\mathbf{0.62\pm0.03}$ & $1.02\pm0.02$ \\
$A2A$         & BB & $0.51\pm0.02$ & $\mathbf{0.11\pm0.01}$ & $\mathbf{0.11\pm0.01}$ & $\mathbf{0.11\pm0.02}$ \\
              & SC & $1.34\pm0.02$ & $1.14\pm0.02$ & $0.43\pm0.01$ & $\mathbf{0.38\pm0.01}$ \\
\bottomrule
\end{tabular}
\caption{RMSD values (in \AA{} units) of reconstructed structures with respect to original atomistic structures. The table presents the average RMSD and standard deviation calculated over the test dataset for each model. The top results for each row are highlighted in bold.}
\label{table:results-proteins}
\end{table*}

The HEroBM framework, powered by GEqTrain, achieves high accuracy and versatility in backmapping diverse molecular systems. The results presented here advance upon the originally published HEroBM methodology \cite{Angioletti2025}, which itself produced strong outcomes. While the original work already leveraged GEqTrain, we subsequently refined the model by incorporating new features. Specifically, the input for each coarse-grained bead was augmented with two scalar features indicating the presence of preceding and succeeding beads in the polymer sequence. This enhancement, easily integrated through GEqTrain's configuration, contributes to the improved results and highlights the framework's utility for iterative model development. Performance was benchmarked against state-of-the-art methods like CG2AT and cg2all, with key results for proteins summarized in Table \textit{\ref{table:results-proteins}}. On the general protein benchmark using the $PDB_{29k}$ test set, HEroBM demonstrated high data efficiency. Despite being trained on the $PDB_{3k}$ subset, which is approximately ten times smaller than the data used for cg2all, HEroBM achieved comparable or improved accuracy for protein backmapping. Both methods yielded Root Mean Square Deviation (RMSD) values below 0.2 \AA{} for backbone atoms (HEroBM: $0.07 \pm 0.02$ \AA{}; cg2all: $0.08 \pm 0.02$ \AA{}) and below 0.5 \AA{} for all heavy atoms (HEroBM: $0.24 \pm 0.03$ \AA{}; cg2all: $0.31 \pm 0.05$ \AA{}), representing the lowest RMSD among the methods compared here. For the challenging task of backmapping intrinsically disordered proteins (IDPs) from the $PED$ dataset using Martini 3.0 mapping, HEroBM consistently outperformed other methods in reconstructing side chain (SC) structures, achieving the lowest RMSD values across all tested PED entries (e.g., $0.54 \pm 0.06$ \AA{} for $PED_{00055}$ SC, compared to $1.22 \pm 0.03$ \AA{} for cg2all and $1.36 \pm 0.03$ \AA{} for CG2AT).

Furthermore, HEroBM showcased strong transfer learning capabilities. A model trained exclusively on a single G protein-coupled receptor (GPCR) system, $A2A$ (denoted $HEroBM_{A2A}$), demonstrated remarkable performance when applied to the $PED$ datasets, outperforming both CG2AT and cg2all in side chain reconstruction and CG2AT for backbone atoms in these distinct systems. For instance, on $PED_{00055}$ side chains, $HEroBM_{A2A}$ achieved $1.00 \pm 0.03$ \AA{}. When trained and tested on $A2A$ itself, $HEroBM_{A2A}$ produced highly accurate backbone ($0.11 \pm 0.02$ \AA{}) and side chain ($0.38 \pm 0.01$ \AA{}) reconstructions.

HEroBM was also evaluated in a more challenging low-information regime, in which the coarse-grained representation retains only the $C_{\alpha}$ atom of each residue. In this setting, the backmapping task becomes substantially more difficult, since the model must reconstruct the full atomistic structure starting from a highly reduced input, with no explicit side-chain information and only limited local structural context. For this application, we used the same dataset construction described in the original HEroBM study, consisting of 49\,553 structures from 78 systems for training, 6\,738 structures from 7 systems for validation, and the PED systems reported in Table~\ref{table:results-ca} for testing.

We compare the $C_{\alpha}$-only HEroBM models with recent generative baselines reported by Han et al.~\cite{Han2024LatentRoad} and Zhang et al.~\cite{Zhang2025Exploit}, who evaluated GenZProt~\cite{Yang2023}, DiAMoNDBack~\cite{Jones2023}, Latent Diffusion Backmapping (LDB) and LatCPB models on the same four PED systems. Their protocol samples each protein structure ten times and reports the mean and standard deviation of the resulting metrics. Because that work reports a single RMSD value per PED system rather than separate backbone and side-chain RMSDs, we list those values only in the ``ALL'' rows of Table~\ref{table:results-ca}. This avoids implying a backbone/side-chain decomposition that was not reported in the original source.

As in the Martini 3.0 application, we augmented the coarse-grained graph with directional information encoding the presence of preceding and succeeding residues along the sequence. This provides an explicit notion of chain directionality, which is otherwise absent in a pure $C_{\alpha}$ trace. The effect of this modification is reported in Table~\ref{table:results-ca}, where the updated model is directly compared with the previous $C_{\alpha}$-only HEroBM version. Directionality consistently improves backbone reconstruction across all four PED test systems, reducing the BB RMSD from $0.62$ to $0.56$~\AA{} on $PED_{55}$, from $0.81$ to $0.69$~\AA{} on $PED_{90}$, from $0.63$ to $0.41$~\AA{} on $PED_{151}$, and from $0.52$ to $0.44$~\AA{} on $PED_{218}$. A similar trend is observed for side chains, with SC RMSD decreasing for all systems.

\begin{table*}[t]
\centering
\caption{C$\alpha$-trace backmapping performance on PED systems. HEroBM values are reported separately for heavy backbone atoms (BB), heavy side-chain atoms (SC), and all heavy atoms (ALL), comparing the previous C$\alpha$-only model with the updated model including directional sequence information. GenZProt, DiAMoNDBack, LDB, and LatCPB report only global RMSD values in the corresponding studies and are therefore included only in the ALL rows. Values for GenZProt, DiAMoNDBack, and LDB are taken from Han et al.~\cite{Han2024LatentRoad}; LatCPB values are taken from Zhang et al.~\cite{Zhang2025Exploit}}
\label{table:results-ca}
\resizebox{\textwidth}{!}{%
\begin{tabular}{@{}llcccccc@{}}
\toprule
Dataset & Atoms & HEroBM (prev.) & HEroBM (+dir.) & GenZProt~\cite{Yang2023} & DiAMoNDBack~\cite{Jones2023} & LDB~\cite{Han2024LatentRoad} & \textbf{LatCPB}~\cite{Zhang2025Exploit} \\
\midrule
$PED_{00055}$ & BB  & $0.62\pm0.06$ & $0.56\pm0.05$ & -- & -- & -- \\
                & SC  & $2.61\pm0.13$ & $2.07\pm0.14$ & -- & -- & -- \\
                & ALL & $1.87\pm0.09$ & $\mathbf{1.51\pm0.10}$ & $1.839\pm0.002$ & $1.843\pm0.008$ & $1.689\pm0.009$ & $1.695$ \\
\addlinespace
$PED_{00090}$ & BB  & $0.81\pm0.05$ & $0.69\pm0.05$ & -- & -- & -- \\
                & SC  & $2.50\pm0.12$ & $2.37\pm0.11$ & -- & -- & -- \\
                & ALL & $1.86\pm0.07$ & $\mathbf{1.73\pm0.07}$ & $2.070\pm0.003$ & $1.958\pm0.014$ & $1.857\pm0.020$ & $1.758$ \\
\addlinespace
$PED_{00151}$ & BB  & $0.63\pm0.09$ & $0.41\pm0.06$ & -- & -- & -- \\
                & SC  & $2.07\pm0.08$ & $1.69\pm0.11$ & -- & -- & -- \\
                & ALL & $1.58\pm0.08$ & $\mathbf{1.25\pm0.07}$ & $1.629\pm0.001$ & $1.769\pm0.008$ & $1.673\pm0.005$ & $1.539$ \\
\addlinespace
$PED_{00218}$ & BB  & $0.52\pm0.03$ & $0.44\pm0.04$ & -- & -- & -- \\
                & SC  & $2.50\pm0.08$ & $2.13\pm0.06$ & -- & -- & -- \\
                & ALL & $1.80\pm0.04$ & $\mathbf{1.53\pm0.04}$ & $1.800\pm0.002$ & $1.637\pm0.012$ & $1.622\pm0.015$ & $1.563$ \\
\bottomrule
\end{tabular}%
}
\end{table*}

Despite the severe reduction in input information, the updated model achieves BB RMSD values between $0.41 \pm 0.06$~\AA{} and $0.69 \pm 0.05$~\AA{}, indicating that the local backbone geometry can still be recovered with high fidelity in the $C_{\alpha}$-only setting. The RMSD computed over all heavy atoms ranges from $1.25 \pm 0.07$~\AA{} to $1.73 \pm 0.07$~\AA{}, confirming that the reconstructed structures remain overall close to the atomistic reference. As expected, the largest contribution to the residual error arises from the side chains, which remain intrinsically more ambiguous to reconstruct in intrinsically disordered proteins when only $C_{\alpha}$ positions are provided.

This comparison also clarifies a methodological distinction between deterministic reconstruction and generative backmapping. Generative models are attractive because a single coarse-grained configuration may correspond to multiple compatible atomistic arrangements. However, diversity is useful only if the generated structures remain consistent with the atomistic conditional distribution associated with the coarse-grained input. In practical multiscale workflows, accurate backmapping and ensemble generation need not be solved by the same model: one can reconstruct high-fidelity atomistic structures from coarse-grained frames and then rely on atomistic relaxation or molecular dynamics to recover local equilibrium fluctuations. The present results therefore suggest that, at least for $C_{\alpha}$-trace protein backmapping under paired-reference evaluation, a one-shot equivariant reconstruction model can outperform current generative alternatives in structural fidelity, while leaving open whether generative sampling provides additional value when evaluated against explicit ensemble-level observables.

We stress that the relevant point here is methodological: the improvement over the originally published HEroBM was obtained by \textbf{adding two scalar connectivity features through configuration alone}, with no change to the underlying equivariant engine, illustrating how the framework supports rapid, low-overhead model refinement.

\subsection{Case Study: NMR Chemical Shift Prediction in Molecular Solids}
\label{sec:nmr}

NMR chemical shifts are local probes of atomic structure: they report how the electronic environment around a nucleus shields it from an external magnetic field. In molecular solids, this makes chemical shifts particularly useful for structure validation and NMR crystallography, where candidate crystal structures can be ranked by comparing experimental shifts with computed ones. The standard computational route relies on first-principles GIPAW-DFT calculations, which are accurate but expensive when many structures or large candidate pools must be screened. Machine learning therefore provides a natural surrogate problem: given an atomistic crystal structure, predict the DFT-computed shielding or shift associated with each atomic environment.

We use this task as a controlled structure--observable benchmark for GEqTrain. Unlike the backmapping task, where the output is a geometry, chemical-shift prediction asks the same equivariant graph stack to produce local scalar observables.

\paragraph{Controlled scalar benchmark on CSD-2k/CSD-500.}
We first evaluate GEqTrain on the original ShiftML benchmark, where models are trained on the \texttt{CSD-2k} set of DFT-relaxed molecular crystals containing H, C, N and O, and evaluated on the held-out \texttt{CSD-500} test set. This setting is useful because it isolates the scalar prediction problem: the model is trained only on isotropic shieldings, without tensorial targets, without pretraining and without ensemble averaging. Results are shown in Table~\ref{tab:csd500_benchmark}.

\begin{threeparttable}[htbp]
\centering
\caption{Prediction accuracies for isotropic chemical shifts/shieldings on the molecular-solid benchmark. Values are RMSEs in ppm. The first block reports models evaluated directly on the original \texttt{CSD-500} benchmark. The second block provides contextual modern-reference values from ShiftML2/3 on a comparable H/C/N/O relaxed subset of the newer CSD-test set.}
\label{tab:csd500_benchmark}
\begin{tabular}{lccccc}
    \toprule
    \textbf{Model} &
    \textbf{$^{1}$H} &
    \textbf{$^{13}$C} &
    \textbf{$^{15}$N} &
    \textbf{$^{17}$O} &
    \textbf{Training setting} \\
    \midrule
    ShiftML1.0 (KRR) & 0.49 & 4.30 & 13.30 & 17.70
    & ShiftML1 only; SOAP/KRR \\
    Unzueta GNN\tnote{a} & 0.49 & 4.06 & 9.90 & 14.40
    & GNN; pooled molecular data \\
    MR-3D-DenseNet & 0.37 & 3.30 & 10.20 & 15.30
    & ShiftML1 only; voxelized 3D grids \\
    GEqTrain & 0.41 & 3.41 & 9.82 & 13.80
    & ShiftML1 only; single model \\
    \midrule
    \textit{NMRNet}\tnote{b} & \textit{0.35} & \textit{3.21} & \textit{9.45} & \textit{13.03}
    & \textit{self-supervised pretraining + fine-tuning} \\
    \textit{ShiftML2}\tnote{c} & \textit{0.47} & \textit{4.07} & \textit{12.52} & \textit{19.50}
    & \textit{larger ShiftML2 pool + thermal distortions} \\
    \textit{ShiftML3}\tnote{c} & \textit{0.39} & \textit{1.97} & \textit{5.71} & \textit{9.91}
    & \textit{tensorial dataset; PET/nanoPET ensemble} \\
    \bottomrule
\end{tabular}

\smallskip
\begin{flushleft}
    \footnotesize
    $^{a}$ Reported GNN model trained on a broader molecular compilation.\\
    $^{b}$ Uses self-supervised geometric pretraining before NMR fine-tuning.\\
    $^{c}$ Not evaluated on the original \texttt{CSD-500}; included as contextual modern-reference values on a comparable H/C/N/O relaxed subset of the newer CSD-test set.
\end{flushleft}
\end{threeparttable}

On the original \texttt{CSD-500} benchmark, GEqTrain remains competitive with methods trained under comparable data conditions. In particular, it is close to MR-3D-DenseNet, improving its errors for $^{15}$N and $^{17}$O while being slightly worse for $^{1}$H and $^{13}$C. It also remains in the same accuracy range as NMRNet, despite NMRNet relying on a substantially more elaborate pretraining-and-fine-tuning strategy. This is the appropriate interpretation of the benchmark: GEqTrain is not presented as a specialized NMR architecture, but as a general equivariant framework that can be retargeted to local scalar observables and reach competitive accuracy from scratch.

The comparison with ShiftML2 and ShiftML3 should be read only as field context. As discussed in the ShiftML3 supplementary material, the original \texttt{CSD-500} and the newer CSD-test set have limited structural overlap, and differences in DFT convergence parameters and pseudopotentials make direct numerical comparison delicate. ShiftML2 and ShiftML3 are therefore shown as modern reference points rather than as strict competitors in the same benchmark protocol.

\paragraph{Extension to the ShiftML3 tensorial dataset.}
We next evaluate the same framework on the newer ShiftML3 dataset, where the target information includes not only isotropic shieldings but also tensorial shielding components. In this setting, the data cover a broader chemical space, include more nuclei, and expose the physical fact that chemical shielding is fundamentally tensorial, even when only its isotropic component is used for many applications.

For this experiment we train a single GEqTrain model over all nuclei, with a shared equivariant trunk and two node-level readouts: one scalar head for \(\sigma_{\mathrm{iso}}\), and one tensorial head for the non-scalar irreducible shielding components \(\sigma_{\mathrm{tensor}}\).

\begin{table}[htbp]
    \centering
    \caption{Isotropic shielding prediction on the ShiftML3 CSD-test split. ShiftML2 and ShiftML3 values are reported on the same CSD-test setting. GEqTrain is trained as a single shared model over all nuclei, without ensemble averaging. Errors are in ppm.}
    \label{tab:shiftml3_iso}
    \begin{tabular}{lcccccc}
        \toprule
        {\textbf{Nucleus}} &
        \multicolumn{2}{c}{\textbf{ShiftML2}} &
        \multicolumn{2}{c}{\textbf{ShiftML3}} &
        \multicolumn{2}{c}{\textbf{GEqTrain single}} \\
        \cmidrule(lr){2-3}
        \cmidrule(lr){4-5}
        \cmidrule(lr){6-7}
        & MAE & RMSE & MAE & RMSE & MAE & RMSE \\
        \midrule
        $^{1}$H  & 0.39 & 0.51 & 0.33 & 0.43 & 0.37 & 0.49 \\
        $^{13}$C & 3.15 & 4.63 & 1.58 & 2.32 & 2.83 & 3.98 \\
        $^{15}$N & 10.29 & 15.72 & 5.00 & 10.41 & 9.86 & 14.78 \\
        $^{17}$O & 16.11 & 22.96 & 7.51 & 11.45 & 15.11 & 21.20 \\
        \bottomrule
    \end{tabular}
\end{table}

On the isotropic component, the single GEqTrain model reaches accuracies close to, and slightly better than, ShiftML2, but remains below ShiftML3. This ordering is expected and is scientifically useful. ShiftML3 is an NMR-specialized model using an ensemble of nanoPET models. GEqTrain, by contrast, is evaluated here as a single shared model using the same configurable infrastructure employed in the other case studies. The result therefore positions GEqTrain as a competitive and reusable framework, rather than as a task-specialized replacement for ShiftML3.

\subsection{Case study III: a deterministic LEGO benchmark for mixed-field equivariant generation}
\label{sec:lego}

As a third case study, we consider a synthetic benchmark designed to test a capability that is not directly exposed by standard coordinate-only generative tasks: the \textbf{joint generation of node positions, higher-order geometric descriptors, and vector-valued attributes} within a single equivariant model.
We stress at the outset that this is a controlled, synthetic benchmark with procedurally generated structures and paired references; it isolates a geometric capability (joint generation of position, local shape and orientation) rather than modeling a physical system. Its role is to test whether the framework can express and train mixed-field equivariant generation, not to demonstrate chemical realism.
The purpose is to probe whether the same typed framework can support generative modeling over heterogeneous node states, including Cartesian coordinates, tensor-like shape descriptors, and dipole-like vectors.

Each node in the benchmark carries three coupled target quantities:
a position $\mathbf{r}_i \in \mathbb{R}^3$, a shape descriptor $\mathbf{s}_i$ represented by a 16-dimensional coefficient vector corresponding to the irreps \texttt{1x0e + 1x1o + 1x2e + 1x3o},
and a dipole-like vector $\boldsymbol{\mu}_i \in \mathbb{R}^3$.
The resulting task therefore extends beyond geometric denoising in Cartesian space and instead requires coherent reconstruction of multiple properties that must remain mutually compatible throughout sampling.

\paragraph{Task:}
The generative problem is formulated as conditional reconstruction of a connected subassembly from its surrounding context.
Given a complete LEGO-like assembly, we partition the structure into a fixed context and a diffused substructure, analogous to the receptor--ligand split used in other conditional generative settings.
The context nodes are kept fixed, while the selected subassembly is corrupted and then reconstructed by the model.
Importantly, corruption and reconstruction are applied jointly to all target fields associated with the diffused nodes, namely positions, shape descriptors, and dipole vectors.
The model must therefore recover not only where the missing bricks should be placed, but also which local shape state they should realize and which dipole orientation they should carry.

In the present implementation, GEqDiff performs direct flow matching on all three channels.
The position head predicts a velocity field for Cartesian coordinates, the shape head predicts a velocity in the 16-dimensional shape space, and the dipole head predicts a velocity in the 3-dimensional dipole space.

\paragraph{Dataset:}
The LEGO dataset is generated procedurally from simple, idealized structural motifs. 
Each structure starts from a scaffold resembling a beta sheet, an alpha helix, or a mixed topology combining the two. 
The scaffold defines the ordered three-dimensional positions of the LEGO elements and their local connectivity. 
Brick identities are then assigned by deterministic rules: \texttt{1x1}, \texttt{1x2} and \texttt{T}-shaped bricks form straight segments, while \texttt{L}-shaped bricks define turns.
Once the brick structure is fixed, local shape descriptors and dipole-like vectors are computed from it and used as additional equivariant fields in the generative task.

This pipeline yields a structured assembly in which each node has a position, a shape state, and a dipole-like vector attribute assigned from local context.
The benchmark is a controlled test of mixed-field generation under equivariance, in which geometry, local shape identity, and directional attributes must be generated consistently.

\begin{figure}[t]
\centering
\includegraphics[width=\textwidth]{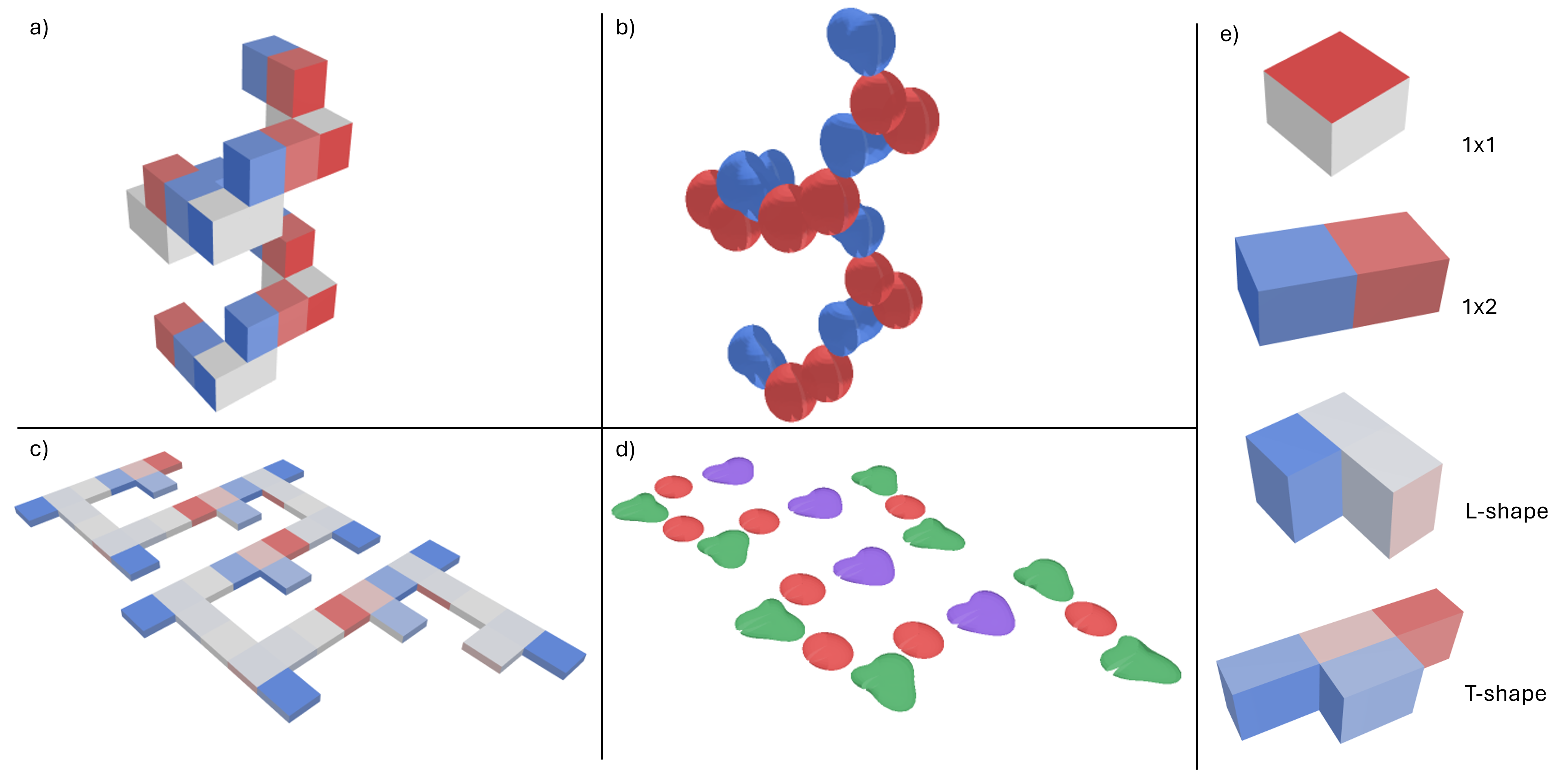}
\caption{
Procedural LEGO benchmark.
(a) Representative alpha-helix-like LEGO assembly.
(b) Spherical-harmonic surface rendering of the shape descriptors associated with the alpha-helix-like structure.
(c) Representative beta-sheet-like LEGO assembly.
(d) Corresponding spherical-harmonic surface rendering.
(e) Primitive brick vocabulary used by the deterministic construction rules, including \texttt{1x1}, \texttt{1x2}, \texttt{L}-shaped, and \texttt{T}-shaped elements.
The surface renderings are used only to visualize the equivariant shape channel; the dataset itself is generated from procedural scaffold and brick-assignment rules.
}
\label{fig:lego_dataset}
\end{figure}

Figure~\ref{fig:lego_dataset} summarizes the construction visually.
Panels (a,c) show representative alpha-helix-like and beta-sheet-like brick assemblies, while panels (b,d) show the corresponding spherical-harmonic surface renderings used to visualize the shape channel.
Panel (e) shows the primitive brick vocabulary used by the deterministic assignment rules.
The smooth surfaces are visualizations of the local shape descriptors.

\paragraph{Results:}\label{lego:results}
The quantitative assessment of the LEGO benchmark is structured to isolate the contribution of each equivariant channel to the overall generative coherence.
The quantitative analysis asks two separate questions:
first, does adding non-coordinate equivariant fields destabilize the generation of a geometrically valid subassembly?
Second, can the same flow-matching model reconstruct the additional shape and vector-valued channels with useful fidelity?
We therefore report metrics for geometry, shape, dipoles, and pose separately.

\begin{figure}[t]
\centering
\includegraphics[width=\textwidth]{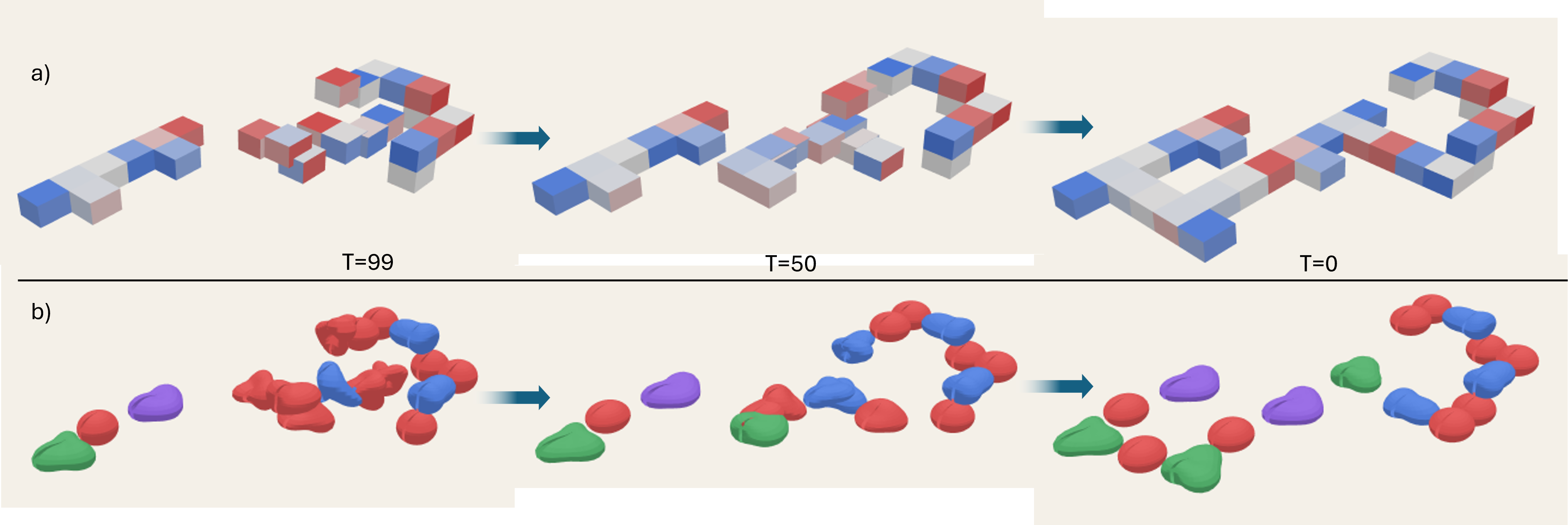}
\caption{
Example mixed-field generation trajectory on the LEGO benchmark.
(a) Brick-level representation of a mixed-topology sample at three stages of the reverse flow, from an initially corrupted state to the final generated assembly.
(b) Corresponding visualization of the predicted spherical-harmonic shape descriptors at the same stages.
The example illustrates the joint reconstruction of coordinates and equivariant shape features during sampling.
}
\label{fig:lego_flow}
\end{figure}

A qualitative sampling trajectory is shown in Figure~\ref{fig:lego_flow} for an example from the mixed scaffold family.
The upper row shows the evolution of the brick-level coordinate representation, while the lower row shows the same sampling stages rendered through the spherical-harmonic shape surfaces.
The visualization illustrates that the model does not only place the diffused subassembly in space, but also reconstructs the associated local shape field during the same flow process.

\textbf{Validity} is a continuous geometric score on a 0--100 scale that penalizes volumetric overlaps, severe brick clashes, and disconnected components.
\textbf{Validity (raw)} is evaluated directly on the continuous decoded coordinates, whereas \textbf{Validity (vox)} is evaluated after projecting the generated anchors onto the discrete LEGO lattice.
Their difference therefore measures the extent to which geometric penalties arise
from sub-lattice placement errors.
\textbf{Shape} is a composite fidelity score combining error in the generated 16-dimensional spherical-harmonic descriptor, decoded brick-type accuracy, and decoded orientation similarity.
\textbf{Dipoles} combines directional and magnitude agreement of the generated dipole-like vectors with the paired reference field.
\textbf{Pose (vox)} is an exponentially weighted score based on the mean and maximum displacement of the generated anchors and the maximum displacement of the fixed context after lattice projection. Exact definitions and calibration constants are provided in the Supplementary Information.

The qualitative trajectory in Figure~\ref{fig:lego_flow} shows the intended behavior of the benchmark: coordinates and shape descriptors are denoised together, rather than generated in separate post-processing steps.
The quantitative results are summarized in Table~\ref{tab:lego_results}.
For the motif-specific full models, lattice projection removes the geometric penalties measured by the benchmark: both the alpha-helix-like and beta-sheet-like regimes obtain maximal voxelized validity and pose scores of 100. Their raw validity scores are lower, at 85.18 and 81.32, respectively, showing that residual overlap or connectivity penalties remain in the continuous decoded coordinates before projection onto the lattice. The non-coordinate channels are also reconstructed consistently, with shape fidelity scores of 90.94 and 89.91 and dipole fidelity scores of 99.44 and 97.84 for the alpha-helix-like and beta-sheet-like regimes, respectively.

\begin{table}[t]
\centering
\caption{
Performance on the LEGO benchmark across structural motifs and diffused field combinations. All entries are scores on a 0--100 scale, with larger values indicating better performance; they should not be interpreted as percentages of samples passing a binary criterion. Validity (raw) is evaluated directly after flow integration and decoding, whereas Validity (vox) is evaluated after projection onto the discrete LEGO lattice. Shape is a composite score combining continuous descriptor fidelity, decoded brick-type accuracy, and orientation similarity. Dipoles measures directional and magnitude fidelity of the generated vector field. Pose (vox) measures anchor preservation relative to the fixed context after lattice projection. Dashes indicate channels that are not generated by the corresponding ablation. Values are means over 1000 test assemblies, with 4 generated samples per assembly.
}
\label{tab:lego_results}
\begin{tabular}{lcccccc}
\toprule
Model variant & Validity (raw) & Validity (vox) & Shape & Dipoles & Pose (vox) \\
\midrule
Alpha: coord+shape+dipole & 85.18 & 100   & 90.94 & 99.44 & 100   \\
Beta:  coord+shape+dipole & 81.32 & 100   & 89.91 & 97.84 & 100   \\
Mixed: coord              & 86.89 & 100   & -     & -     & 100   \\
Mixed: shape              & -     & -     & 92.30 & -     & -     \\
Mixed: dipole             & -     & -     & -     & 98.45 & -     \\
Mixed: coord+shape+dipole & 83.25 & 99.00 & 91.87 & 98.45 & 99.00 \\
\bottomrule
\end{tabular}
\end{table}

The mixed benchmark is the more informative setting because it combines heterogeneous scaffold grammars.
A coordinate-only model obtains raw and voxelized validity scores of 86.89 and 100, respectively, confirming that the positional reconstruction problem is learnable in isolation. The single-channel models obtain a shape fidelity score of 92.30 and a dipole fidelity score of 98.45. Most importantly, the full mixed-field model jointly transports coordinates, shape descriptors, and dipole-like vectors while retaining near-maximal voxelized validity and pose scores of 99.00. Its shape and dipole scores, 91.87 and 98.45, remain close to those of the corresponding single-channel models. The lower raw validity score of 83.25 shows that joint generation does not remove continuous off-lattice placement errors; however, most of the associated geometric penalty disappears after projection onto the discrete lattice.

In conclusion, GEqDiff can formulate and train a single flow-matching model over coordinates together with higher-order equivariant shape descriptors and vector-valued attributes, without catastrophic interference between the generated fields.
The gap between raw and voxelized validity further indicates that many apparent geometric failures are small off-lattice inaccuracies rather than failures of the underlying scaffold grammar.

\section{Discussion}

A recurring bottleneck in equivariant deep learning is not architectural but organizational: adapting an existing EGNN implementation to a new dataset, feature space, training protocol, or scientific question typically requires non-trivial re-engineering of task-specific code.
GEqTrain addresses this by treating dataset semantics, geometric representations, and training objectives as separable configuration layers, so that models, attributes, losses, and workflows can be composed from reusable building blocks rather than re-implemented for each application.
The main contribution of GEqTrain is therefore \textit{organizational}: it lowers the cost of moving between tasks, output types and training regimes by making them configuration choices over a shared equivariant stack. It is complementary to, not competitive with, highly optimized single-purpose systems (e.g. dedicated backmapping pipelines, pre-trained chemical-shift models, or specialized 3D generators), which remain the right tools when maximal per-task accuracy is the only objective.
Second, within this configurable interface, GEqDiff extends the \textit{target space} of equivariant generation from coordinates-plus-scalars to coupled higher-order equivariant fields, a novel capability that becomes scientifically relevant when geometry must be generated together with orientation, local shape, or polarization-like attributes.
The three case studies reported here test whether this design holds up across settings that differ substantially in target type and training regime.

The NMR results show that the same general-purpose stack can be applied to scalar regression in a data-limited molecular benchmark.
HEroBM provides a more demanding structural test.
Hierarchical backmapping requires mapping single coarse-grained beads to complex, multi-atom geometries, which inherently requires staged coordinate reconstruction.
GEqTrain separates the equivariant feature extraction from the task-specific topology, using a dedicated reconstruction module appended to the pipeline.
This demonstrates how complex, domain-specific structural generation can be integrated as a task layer without changing the underlying equivariant engine.

The results on the LEGO benchmark show that the generative target space can be extended beyond Cartesian coordinates to include higher-order shape descriptors and vector-valued attributes.
The full mixed-field model maintains 99\% voxelized validity and pose preservation on the heterogeneous mixed benchmark while reconstructing the shape and dipole channels with accuracies comparable to the corresponding single-channel ablations.
The strict raw validity is lower, which exposes small continuous off-lattice inaccuracies before voxelization, but the near-perfect voxelized validity indicates that the generated samples usually preserve the intended discrete scaffold after projection to the lattice.
Thus, the main message is that auxiliary equivariant fields can be included as first-class generative targets in the equivariant flow-matching process.

Many scientific generative problems are not naturally coordinate-only: one may want to generate positions together with local shape descriptors, orientation fields, polarization-like vectors, pharmacophore occupancy, or other tensorial quantities coupled to geometry.
The LEGO task demonstrates that such mixed-field objectives can be expressed within the typed GEqTrain/GEqDiff interface when the corresponding fields are provided as model inputs and outputs.
However, transfer to physically meaningful systems remains an open problem.
The current benchmark uses procedural rules and paired references, and therefore does not test thermodynamic weighting, chemical validity, long-range physical consistency, or multimodal sampling over realistic molecular configurations.

Two design features are worth making explicit.
First, the current interaction backbone emphasizes local message passing, which supports memory-bounded chunked execution for large node- and edge-level problems.
This extends the accessible system size under a fixed memory budget, although serial
chunk construction and evaluation increase inference time (Supplementary Fig. 1).

At the same time, tasks dominated by long-range couplings or global context will likely benefit from extensions based on global pooling, cross-scale communication, or hybrid attention mechanisms, which are supported within the GEqTrain framework and depend on the architecture of choice.
Second, the Hydra-based configuration system provides a clear route to reproducibility, systematic ablation, and structured reuse across experiments.

As a future research direction, we plan to put pressure on the framework's organizing principle through multi-target settings with jointly regressed scalar and equivariant outputs, larger heterogeneous datasets, and generative tasks with structured physicochemical conditioning.
Examples include protein-pocket shape, pharmacophore occupancy, ligand orientation, and polarization-like vector fields, where the geometric vocabulary introduced in the LEGO benchmark becomes directly relevant but must be evaluated under realistic physical constraints.

In conclusion, GEqTrain aims to show that a typed, configuration-driven interface over data semantics, geometry, and training objectives reduces the cost of moving between scalar prediction, structural reconstruction, and mixed-field generative modeling.
The present case studies support this organizational principle while also clarifying its current limits: the framework makes such experiments easier to define and ablate, but scientific validity still depends on the quality of the data, targets, and evaluation protocols used for each domain.

\section{Code, Data, and Reproducibility}

GEqTrain is available under the MIT license at \url{https://github.com/limresgrp/GEqTrain}.
The repository contains the training and inference framework, the Hydra configuration files defining the experiments, environment-setup utilities, automated tests, and tutorials for constructing datasets and training scalar and tensorial equivariant models.
In particular, the chemical-shift tutorial provides the preprocessing scripts, dataset definitions, and experiment, data, model, and training configurations used for the molecular-solid NMR workflow.

The GEqDiff generative extension is available under the MIT license at \url{https://github.com/limresgrp/GEqDiff}.
It contains the complete LEGO workflow, including procedural dataset generation, construction of masked flow-matching datasets, training configurations, sampling, quantitative evaluation, and visualization scripts.

The HEroBM backmapping implementation is available under the MIT license at
\url{https://github.com/limresgrp/HEroBM}.
The repository includes the configuration and preprocessing infrastructure used for backmapping, deployed model support, an end-to-end tutorial, and detailed command-line
documentation.


\printbibliography
\end{document}


\title{Supplementary Information for \\ GEqTrain: A Configuration-Driven Framework for Retargeting Equivariant Graph Neural Networks Across 3D Scientific Tasks}

\author[1]{Daniele Angioletti}
\author[1]{Marco Nobile}
\author[1]{Vittorio Limongelli\thanks{Correspondence to: \href{mailto:vittoriolimongelli@gmail.com}{vittoriolimongelli@gmail.com}}}

\affil[1]{Faculty of Biomedical Sciences, Euler Institute, Universit\'a della Svizzera italiana, 6900 Lugano, Switzerland}

\maketitle

\section{Irreducible representations and spherical harmonics in \texttt{e3nn}}
\label{sec:supp_irreps}

GEqTrain uses the \texttt{e3nn} convention for irreducible representations (\emph{irreps}) of \(O(3)\) \cite{e3nn,e3nn_software}. An irrep is indexed by an angular order \(l\) and a parity \(p\in\{+1,-1\}\), written in the code as \texttt{e} (even) or \texttt{o} (odd). The corresponding feature block has dimension \(2l+1\), so that \(l=0\) yields one component, \(l=1\) yields three components, and \(l=2\) yields five components. Direct sums of such blocks are written as strings such as \texttt{64x0e} for representing a tensor of 64 scalar invariant quantities or \texttt{1x1o + 1x2e + 1x3o} for heterogeneous tensor composed of higher-order equivariant channels.

Spherical harmonics provide a canonical way to construct such angular channels from relative directions. For an edge direction \(\hat{\mathbf{r}}_{ij}\), the real spherical harmonics \(Y^l(\hat{\mathbf{r}}_{ij})\) form an equivariant basis of dimension \(2l+1\), satisfying
\[
Y^l(R\hat{\mathbf{r}}) = D^l(R)\,Y^l(\hat{\mathbf{r}}),
\]
which makes them a natural choice for angular edge features in equivariant graph networks. In GEqTrain, these features are computed by \texttt{SphericalHarmonicEdgeAngularAttrs}, while radial distance information is encoded separately through invariant basis expansions (e.g. by projecting onto Gaussian or Bessel functions).

\section{Hydra Configuration for Scalar NMR Prediction}
\label{sec:supp_shiftml3_config}

The current \texttt{GEqTrain} framework uses Hydra composition to assemble an experiment from separate \texttt{/data}, \texttt{/model}, and \texttt{/train} configuration groups (see Listing~\ref{lst:supp_cs_exp}). To provide a concrete example aligned with the scalar NMR task discussed in the main text, we report here a compact configuration for predicting (scalar) chemical shifts on organic crystals in periodic boundary conditions.
The full repository additionally supports richer multitask variants, including joint prediction of scalar and tensorial observables, but these extensions are not required for the scalar-only results discussed in the present study. The content of Listing \ref{lst:supp_cs_exp} can be saved in a file experiment.yaml (an arbitrary name) such to then execute the experiment.

\begin{listing}[h]
\centering
\caption{Compact Hydra experiment composition for scalar chemical-shift prediction.}
\label{lst:supp_cs_exp}
\begin{lstlisting}[language=yaml,gobble=4]
    defaults:
      - /base
      - /data: cs_scalar
      - /model: cs_scalar_local
      - /train: cs_scalar
      - _self_

    root: /path/to/myproj
    run_name: cs_scalar_local
\end{lstlisting}
\end{listing}

\paragraph{Packed NPZ datasets and masking for variable-size examples.}

\label{sec:supp_masked_npz}

The dataset interface shown in Listing~\ref{lst:supp_cs_data} illustrates the simplest case, in which input fields such as \texttt{pos} and \texttt{node\_types} and the target field \texttt{cs\_iso} are read from an NPZ source and mapped to internal GEqTrain field names through \texttt{key\_mapping}. In practice, GEqTrain supports two complementary NPZ-based storage patterns: a dataset may be represented as a collection of NPZ files, for example one file per molecular system, with each file containing one or more frames. In this regime, quantities that are shared across all frames of a file can be declared as \texttt{fixed\_fields} (e.g., atom types usually remain the same from frame to frame). \noindent Alternatively, for large datasets it is often more efficient to pack many examples into a single NPZ archive. When the number of atoms varies across examples, arrays are typically padded to a common size and accompanied by boolean mask fields whose names follow the convention \texttt{<field>\_\_mask\_\_}. These masks specify which rows correspond to valid entries and which correspond to padding. In this regime, fields that were previously constant within a per-system file may instead be stored as frame-dependent padded arrays, and are therefore declared as ordinary node fields rather than fixed fields.

A minimal masked configuration is shown in Listing~\ref{lst:supp_cs_data_masked}. Here, \texttt{pos}, \texttt{node\_types}, and \texttt{cs\_iso} are each paired with a corresponding mask field. During dataset processing, the masks are used to discard padded rows before constructing the internal data object.
In the packed representation, \texttt{node\_types} is no longer stored once per system as a frame-invariant array, but as a padded per-example tensor aligned with \texttt{pos}. The accompanying mask field \texttt{node\_types\_\_mask\_\_} ensures that only the valid rows are retained.

\begin{listing}[h]
\centering
\caption{Scalar-only dataset configuration pointing to multiple npz dataset files in a folder, each representing a single system, potentially with multiple frames.}
\label{lst:supp_cs_data}
\begin{lstlisting}[language=yaml,gobble=4]
    train_dataset_list:
      - dataset: npz
        dataset_input: /path/to/train/folder
        key_mapping:
          pos: pos
          atom_types: node_types
          cs_iso: cs_iso
          Lattice: cell

    validation_dataset_list:
      - dataset: npz
        dataset_input: /path/to/valid/folder
        key_mapping:
          pos: pos
          atom_types: node_types
          cs_iso: cs_iso
          Lattice: cell

    node_fields:
      - cs_iso

    fixed_fields:
      - node_types

    normalization:
      cs_iso:
        mode: per_type:1x0e
        transform: yeo_johnson

    num_types: 21 # Atoms from H to Ca, including unknown atom type X
    avg_num_neighbors: 30
\end{lstlisting}
\end{listing}

\begin{listing}[h]
\centering
\caption{Scalar-only dataset configuration using a single masked npz containing multiple systems, padded and masked.}
\label{lst:supp_cs_data_masked}
\begin{lstlisting}[language=yaml,gobble=4]
    train_dataset_list:
      - dataset: npz
        dataset_input: /path/to/train/masked.npz
        key_mapping:
          pos: pos
          pos__mask__: pos__mask__
          atom_types: node_types
          atom_types__mask__: node_types__mask__
          cs_iso: cs_iso
          cs_iso__mask__: cs_iso__mask__

    node_fields:
      - node_types
      - cs_iso
    # No need to map __mask__ fields
    # they inherit their corresponding field type (node, edge, graph)
\end{lstlisting}
\end{listing}

\paragraph{Normalization}
GEqTrain also supports dataset-level normalization of selected input or target fields through the \texttt{normalization} block. Each field is first optionally transformed by a monotonic map, such as \texttt{signed\_log1p} or \texttt{yeo\_johnson} \cite{Yeo2000ANF}, and then standardized either globally or per node type. Importantly, the transformation parameters and standardization statistics, e.g. means and standard deviations, are fitted only on the training dataset. The same training-fitted parameters are then reused to standardize validation, test, and inference inputs, avoiding split-specific rescaling and preventing information leakage from evaluation data. In the scalar NMR example above, \texttt{mode: per\_type:1x0e} indicates that the isotropic chemical shift is treated as a scalar field and standardized separately for each atom type.

During training, losses are evaluated on the transformed and normalized targets.
During evaluation, GEqTrain applies the corresponding inverse transformation and inverse standardization when computing standard regression-style metrics, so that reported errors such as MAE are expressed again in the original physical units.
For equivariant fields, the current implementation applies mean centering only to scalar irreps, while higher-order components are scaled but not shifted, preserving invertibility of the representation.

\begin{listing}[h]
\caption{Minimal training objective and evaluation block for scalar chemical shift prediction.}
\label{lst:supp_cs_train_metrics}
\begin{lstlisting}[language=yaml,gobble=4]
    loss_coeffs:
      - cs_iso:
        - 1.0
        - geqtrain.train.LogCoshLoss

    metrics_components:
      - cs_iso:
        - L1Loss
        - PerSpecies: true # Show marginal on each atom type

    metrics_key: validation_loss
    metric_criteria: decreasing

    batch_size: 8
    validation_batch_size: 16
    max_epochs: 10000
    learning_rate: 1.e-4
\end{lstlisting}
\end{listing}

\paragraph{Training objective and evaluation setup.}
Listing~\ref{lst:supp_cs_train_metrics} shows a compact optimization and evaluation setup for the scalar NMR task.
Training objectives in GEqTrain are defined as weighted sums of field-specific loss components. If \(\hat{y}_{m}\) denotes the prediction associated with target field \(m\), the total objective takes the generic form
\[
\mathcal{L}=\sum_m \lambda_m \,\mathcal{L}_m\!\left(\hat{y}_{m}, y_{m}\right),
\]
where both the weights \(\lambda_m\) and the underlying loss functionals are declared in the configuration.
This makes it possible to combine several loss terms on the same target field, or to mix node-, edge-, and graph-level supervision within a single training run.
Evaluation metrics are configured separately from the training objective and may therefore use different functionals, such as in the example, reporting \texttt{L1Loss} even when training with a Log-Cosh regression loss.
By default, NaN values are treated as invalid targets: non-finite entries propagate through the loss or metric computation and cause the run to fail. For datasets with intentionally missing or undefined labels, this behavior can be changed for the corresponding loss or metric by setting \texttt{ignore\_nan: true}, in which case only finite prediction--target pairs contribute to the reported value.
When relevant, metrics can also be aggregated per species through \texttt{PerSpecies: true}.
The quantity used for model selection is specified by \texttt{metrics\_key}. In the example Listing~\ref{lst:supp_cs_train_metrics}, \texttt{validation\_loss} selects the checkpoint with the best validation loss. The direction of improvement is controlled by \texttt{metric\_criteria}: \texttt{decreasing} is appropriate for losses and errors, whereas \texttt{increasing} is used for scores where larger values are better, such as accuracy or AUROC.
We show only the subset of optimization hyperparameters needed to interpret the example; the remaining optimizer and scheduler settings are standard training controls and are omitted here for brevity.

\paragraph{Configuring the model stack.}
Listing~\ref{lst:supp_cs_model} illustrates how a GEqTrain model is specified declaratively in configuration space. Rather than constructing a monolithic architecture in code, the model is defined as an ordered \texttt{stack} of modules, each operating on named internal fields and passing its outputs to subsequent stages. In this example, the first entries reuse common \texttt{stack\_blocks} presets for node input attributes and for radial and angular edge geometry, after which task-specific modules are instantiated explicitly through their \texttt{\_target\_} declarations. The stack then applies attribute embedding, local equivariant interaction, edge-to-node reduction, and a final readout head producing the target field \texttt{cs\_iso} with output irreps \texttt{1x0e}.

\begin{listing}[h]
\caption{Scalar-target local model stack for NMR chemical-shift prediction.}
\label{lst:supp_cs_model}
\begin{lstlisting}[language=yaml,gobble=4]
    defaults:
      - /model/stack_blocks: common
      - _self_

    model:
      stack:
        - ${stack_blocks.node_input_attrs}
        - ${stack_blocks.edge_radial_attrs}
        - ${stack_blocks.edge_angular_attrs}
        - _target_: geqtrain.nn.EmbeddingAttrs
          name: attrs
          node_out_irreps: 64x0e
          edge_out_irreps: 64x0e
          edge_eq_out_irreps: 8x1o+8x2e
        - _target_: geqtrain.nn.InteractionModule
          num_layers: 2
          latent_dim: 128
          eq_latent_multiplicity: 8
          output_ls: [0,1,2]
        - ${stack_blocks.edge_pooling}
        - _target_: geqtrain.nn.ReadoutModule
          name: head_cs
          field: node_features
          out_field: cs_iso
          out_irreps: 1x0e
\end{lstlisting}
\end{listing}

\noindent This example also illustrates an important feature of GEqTrain: a scalar observable does not require a purely scalar internal representation. Although the final readout is scalar, the intermediate embedding and interaction stages include both invariant and equivariant channels, allowing the model to remain symmetry-aware throughout the computation while matching the physical type of the observable at the output level.

For readability, the listing relies on reusable \texttt{stack\_blocks} presets for common processing stages. Later sections unpack the main module types underlying this stack and describe in more detail how interaction, reduction, and readout are implemented.

\paragraph{Filtering Nodes for Training}\label{sec:filter_nodes}

GEqTrain provides built-in functionality for selectively filtering nodes based on their type, enabling users to specialize models for specific atom categories without modifying the underlying dataset.
This is achieved using the special keyword \texttt{node\_types}, along with two configuration options:
\begin{itemize}
    \item \texttt{keep\_type\_names}: specifies a list of node types (using regular expressions) to retain for training.
    \item \texttt{exclude\_type\_names\_from\_edges}: specifies node types whose edges to neighboring nodes should be removed during graph construction.
\end{itemize}

\noindent This mechanism allows, for example, training a model to predict properties only for heavy atoms, without needing to manually create multiple `.npz` datasets for each subset of atoms.
An example YAML configuration demonstrating node filtering is shown in Listing~\ref{lst:nodefiltering}.

\begin{listing}[h]
    \caption{Example YAML configuration for filtering nodes and edges based on atom types.}
    \label{lst:nodefiltering}
    \begin{lstlisting}[language=yaml,gobble=4]
    type_names: [X, H, He, Li, Be, B, C, N, O, F]

    # --- Node and edge filtering --- #
    keep_type_names: [H, C, N, O]
    exclude_type_names_from_edges:
      - H # Exclude edges between central atoms and neighboring H atoms
    \end{lstlisting}
\end{listing}

In this example:
\begin{itemize}
    \item \texttt{type\_names} assigns a human-readable label to each node type index, where the number of entries must match the \texttt{num\_types} parameter defined under \texttt{node\_attributes: node\_types}.
    \item \texttt{keep\_type\_names} selects nodes to keep when building dataset (dropping all nodes not corresponding to any of those)
    \item \texttt{exclude\_type\_names\_from\_edges} ensures that, when constructing the local graph around a selected atom (e.g., Carbon), edges connecting to neighboring atoms of type Hydrogen are ignored, effectively limiting the graph neighborhood to heavy atoms only.
\end{itemize}

\noindent This filtering mechanism is particularly useful for tasks such as:
\begin{itemize}
    \item Specialized prediction (e.g., chemical shifts only for one atomic type).
    \item Simplifying the local chemical environment during message passing.
    \item Reducing computational complexity by pruning unnecessary nodes and edges without modifying the underlying original data file.
\end{itemize}

\noindent By leveraging regular expressions for node selection and edge exclusion, GEqTrain allows users to adaptively define multiple training regimes from a single dataset, promoting efficient experimentation without extensive data preprocessing.

\section{\texttt{GEqTrain} Model Architecture} \label{sec:model_architecture}

\subsection*{Molecular graph representation and notation}
In GEqTrain, each sample is represented as a directed geometric graph
\[
\mathcal G=(\mathcal V,\mathcal E),
\]
whose nodes correspond to atoms or coarse-grained sites and whose edges are constructed from a local neighborhood criterion. Given positions \(\mathbf r_i\in\mathbb R^3\), the directed edge set is
\[
\mathcal E=\{(i,j): i\neq j,\ \|\mathbf r_{ij}\|\le r_{\max}\},
\qquad
\mathbf r_{ij}=\mathbf r_j-\mathbf r_i.
\]
For periodic systems, \(\mathbf r_{ij}\) is understood as the lattice-aware relative displacement produced by neighbor construction.

As in the main text, we distinguish three levels of representation. Raw typed inputs are denoted by \(x_i\), \(x_{ij}\), and \(x_G\) for node-, edge-, and graph-level quantities. Their embedded counterparts are denoted by \(a_i\), \(a_{ij}\), and \(a_G\). Geometry-derived edge attributes are written separately as an invariant radial embedding \(\rho_{ij}\) and an equivariant angular embedding \(\mathbf y_{ij}^{\mathrm{sh}}\). This notation is intended to separate \emph{what a quantity means} from \emph{how it is later processed} inside the interaction stack.

This distinction is particularly useful in GEqTrain, where the same model skeleton may be reused across tasks with different data semantics. For example, atom types, NMR chemical shifts, graph-level energies, and externally supplied equivariant descriptors are all introduced as typed fields through configuration, even though they play different physical roles in the final model.

\subsection*{Typed attribute embedding}
The first learned stage of the stack converts user-provided typed inputs into internal attribute fields that can be consumed by subsequent equivariant modules.

\paragraph{\texttt{EmbeddingInputAttrs}:} maps raw typed attributes to invariant and, when present, equivariant internal tensors. For a categorical input field \(f\) associated with node \(i\), the module applies either one-hot encoding or a learned embedding,
\[
\phi_f(x_i^f)=
\begin{cases}
\mathrm{onehot}(x_i^f), & \text{if one-hot encoding is requested},\\[4pt]
\mathbf E_f[x_i^f], & \text{if a learnable embedding table is used},
\end{cases}
\]
while continuous invariant inputs are passed through directly after type registration. The resulting invariant node input attributes are concatenated into a single field,
\[
a_i^{\mathrm{in}}=\big\|_{f\in\mathcal F_{\mathrm{inv}}}\phi_f(x_i^f),
\]
and equivariant inputs, if present, are concatenated analogously into an equivariant field \(a_{i,\mathrm{eq}}^{\mathrm{in}}\). The same logic applies to edge- and graph-level typed inputs.

Conceptually, this module performs the transition from dataset semantics to model-ready typed tensors. It does not yet combine user-defined attributes with geometry; rather, it prepares them in a form that can later be fused with geometric information in a representation-consistent way.

\paragraph{\texttt{EmbeddingAttrs}:} consolidates the available invariant and equivariant typed inputs into the canonical internal attribute fields used by the rest of the stack, such as \texttt{node\_attrs}, \texttt{node\_eq\_attrs}, \texttt{edge\_attrs}, and \texttt{edge\_eq\_attrs}. In practice, this module acts as the bridge between raw user-facing typed inputs and the internal representation expected by interaction blocks and readout modules.

In the local stacks considered in this work, the most common pattern is that node input attributes are embedded first, while geometry-derived edge fields are constructed in parallel and then made available to the interaction backbone. More general tasks may additionally include explicit edge- or graph-level typed input attributes through the same configuration mechanism.

\subsection*{Geometry-derived edge encodings}
In addition to user-defined typed attributes, GEqTrain constructs edge-local geometric descriptors directly from the molecular graph. These geometry-derived fields provide the metric and angular context required for equivariant local interaction.

For each directed edge \((i,j)\), we define
\[
d_{ij}=\|\mathbf r_{ij}\|,
\qquad
\hat{\mathbf r}_{ij}=\frac{\mathbf r_{ij}}{d_{ij}}.
\]

\paragraph{\texttt{BasisEdgeRadialAttrs}:} maps the scalar edge distance \(d_{ij}\) to an invariant radial embedding
\[
\rho_{ij}=\phi(d_{ij}),
\]
where \(\phi\) denotes a radial basis expansion, optionally modulated by a cutoff envelope. This field carries purely invariant metric information and is used to inform scalar latent channels or to gate equivariant interactions.

\paragraph{\texttt{SphericalHarmonicEdgeAngularAttrs}:} maps the unit direction \(\hat{\mathbf r}_{ij}\) to an equivariant angular embedding expressed in spherical harmonics,
\[
\mathbf y_{ij}^{\mathrm{sh}}
=
\bigoplus_{l\le l_{\max}} Y^{(l)}(\hat{\mathbf r}_{ij}).
\]
This construction supplies the directional information required to build higher-order equivariant channels. In contrast to the radial embedding, which is invariant, \(\mathbf y_{ij}^{\mathrm{sh}}\) transforms according to the irreducible representations selected by the truncation order \(l_{\max}\).

Together, \(\rho_{ij}\) and \(\mathbf y_{ij}^{\mathrm{sh}}\) define the geometry-derived edge context used throughout the local interaction stack. They are kept conceptually distinct from user-defined edge attributes because they arise deterministically from positions and graph construction, rather than from external annotations or dataset-specific metadata.

\subsection*{Scalar and equivariant primitives}

The local interaction stack in GEqTrain is built from a small set of reusable low-level primitives that separate invariant processing from equivariant processing while preserving a common interface. Throughout this section, we follow the conventions introduced in the main text and in the preceding supplementary sections: scalar features are denoted by \(s\), equivariant features by \(\mathbf q\), and their associated transformation types are declared explicitly through \(O(3)\) irreps. The role of the primitives is then as follows: \texttt{ScalarMLPFunction} processes invariant channels only, \texttt{SO3\_Linear} applies equivariant linear mixing within matched irrep types, \texttt{SO3\_LayerNorm} normalizes irreps tensors in a representation-aware manner, and \texttt{EquivariantScalarMLP} combines these ingredients into a hybrid scalar--equivariant processing block. 

\paragraph{\texttt{ScalarMLPFunction}:}
is the basic fully connected block for invariant channels. Given an input \(s\in\mathbb R^{d_{\mathrm{in}}}\), it constructs a standard multilayer perceptron with user-configurable hidden dimensions, nonlinearity, optional dropout, optional weight normalization, and an optional \texttt{LayerNorm} inserted before the first linear layer \cite{Xiong2020LayerNorm}. If the chosen nonlinearity is \texttt{swiglu}, intermediate widths are doubled before gating; if \texttt{zero\_init\_last\_layer\_weights} is enabled, the last linear layer is initialized with a reduced scale, such to make the last layer of the \texttt{ScalarMLPFunction} output values around zero. Thus, in abstract form,
\[
\phi_{\mathrm{MLP}}(s)
=
L_K\circ \sigma_{K-1}\circ L_{K-1}\circ \cdots \circ \sigma_1 \circ L_1 \circ \mathrm{Norm}(s),
\]
where the exact choice of normalization, nonlinearity, dropout, and bias is configuration-dependent. This module is used whenever GEqTrain must process invariant latent channels or generate scalar-dependent modulation weights. 

\paragraph{\texttt{SO3\_Linear}:}
implements a fully connected \(SO(3)/O(3)\)-equivariant linear map that mixes only \emph{multiplicities} within matching irrep types, i.e.\ between channels with the same angular order \(l\) and parity. If
\[
\mathbf q = \bigoplus_{\alpha} \mathbf q^{(\alpha)}, 
\qquad
\mathbf q^{(\alpha)} \in \mathbb R^{m_\alpha^{\mathrm{in}}\times (2l_\alpha+1)},
\]
and the output contains the same irrep type with multiplicity \(m_\alpha^{\mathrm{out}}\), then the action of the layer on that block is
\[
\widetilde{\mathbf q}^{(\alpha)}
=
W^{(\alpha)} \mathbf q^{(\alpha)},
\qquad
W^{(\alpha)}\in\mathbb R^{m_\alpha^{\mathrm{out}}\times m_\alpha^{\mathrm{in}}},
\]
with the \((2l_\alpha+1)\)-dimensional irrep axis left untouched. In the implementation this map supports both flat tensors of shape \((B,\mathrm{dim})\) and channel-wise tensors of shape \((B,m,\mathrm{dim}/m)\) when a common multiplicity exists. It can use internal learnable weights or externally supplied packed weights, and bias is added only to scalar output blocks (\(l=0\)).

\paragraph{\texttt{SO3\_LayerNorm}:}
provides irreps-aware normalization for tensors whose irreps share a common multiplicity. The implementation groups contiguous blocks with the same angular order \(l\), reshapes them to separate the irrep dimension from the multiplicity dimension, computes a single batch-wise normalization factor per contiguous \(l\)-group, rescales the group, and then restores the original layout. Three normalization modes are supported: \texttt{norm}, which uses the sum of squared components over the irrep dimension; \texttt{component}, which uses their mean; and \texttt{std}, which is identical to \texttt{component} but includes an additional factor \(1/\sqrt{N_l}\), where \(N_l\) is the number of distinct angular orders present, to improve stability with depth. Scalar bias, when enabled, is added only to \(l=0\) groups and is stored in channel-wise form. Denoting a contiguous \(l\)-group by \(\mathbf q_g\), the operation can be summarized as
\[
\widetilde{\mathbf q}_g
=
\frac{\mathbf q_g}{\sqrt{\mathrm{stat}(\mathbf q_g)+\varepsilon}}\cdot \lambda_l
\;+\;
\mathbf b_g^{(0)},
\]
where \(\mathrm{stat}\) depends on the selected normalization mode, \(\lambda_l=1\) for \texttt{norm}/\texttt{component} and \(\lambda_l=1/\sqrt{N_l}\) for \texttt{std}, and \(\mathbf b_g^{(0)}\) is nonzero only for scalar groups. 

\paragraph{\texttt{EquivariantScalarMLP}:}
is the main hybrid primitive used throughout the interaction stack. It accepts either a single tensor containing both scalar and equivariant channels, or an explicit split input \((s,\mathbf q)\). Likewise, it can return either a concatenated output or a split pair of scalar and equivariant outputs. Internally, the module separates scalar irreps from higher-order equivariant irreps and processes the two branches with coupled but distinct mechanisms.

If scalar input channels are present, they define the invariant stream. When an additional scalar conditioning tensor \(c\) is provided, the input scalars are first modulated through FiLM,
\[
s_c =
\begin{cases}
\mathrm{FiLM}(s;c), & \text{if conditioning is provided},\\
s, & \text{otherwise},
\end{cases}
\qquad
\mathrm{FiLM}(s;c)=\gamma(c)\odot s+\beta(c).
\]
The scalar output is then obtained as
\[
\widetilde{s}=\phi_{\mathrm{MLP}}(s_c).
\]
Thus, when both \(s\) and \(c\) are available, conditioning acts by modifying the scalar input representation before the scalar MLP.

The equivariant branch applies an equivariant linear map to the equivariant features \(\mathbf q\). By default, an Allegro-style \cite{Musaelian2023} \texttt{Linear} is used when the input and output irreps have compatible common multiplicities; otherwise, the implementation falls back to \texttt{SO3\_Linear}, which is less optimized but handles mixed multiplicities. The weights of this equivariant map are chosen according to the available invariant information:
\[
\widetilde{\mathbf q} =
\begin{cases}
L_{\mathrm{int}}(\mathbf q), 
& \text{if no scalar input and no conditioning are used},\\[4pt]
L_{\mathrm{ext}}\!\left(\mathbf q;\psi(s_c)\right),
& \text{if scalar input channels are present},\\[4pt]
L_{\mathrm{ext}}\!\left(\mathbf q;\psi(c)\right),
& \text{if no scalar input is present but conditioning is available}.
\end{cases}
\]
Importantly, in the second case the equivariant weights are generated from the same conditioned scalar representation \(s_c\) that feeds the scalar MLP, not from the scalar output \(\widetilde{s}\). Therefore, \(\widetilde{s}\) is the scalar output branch, while \(s_c\) is the invariant control signal used to parameterize the equivariant linear readout.

The module also supports channel-wise output formatting when the multiplicity structure permits it. In split-output mode, if the requested equivariant output itself contains scalar irreps, the scalar input channels can be merged back into the equivariant readout input so that those scalar components are produced consistently by the equivariant branch. In this sense, \texttt{EquivariantScalarMLP} acts as a configurable bridge between invariant conditioning, scalar latent, and equivariant feature processing.

\paragraph{Role in the local interaction stack.}
These primitives reappear repeatedly in the higher-level modules discussed below. In particular, \texttt{EquivariantScalarMLP} is used to generate the initial latent state from concatenated scalar and equivariant edge inputs, to embed edge contributions into local environments, to project tensor-product outputs back to mixed scalar--equivariant latent states, and to perform the final projection to the requested output irreps. \texttt{SO3\_LayerNorm} is applied immediately after tensor-product contraction in each interaction layer, and \texttt{SO3\_Linear} is used explicitly in the initial and final latent projections whenever an equivariant linear map with externally generated or mixed-multiplicity weights is required. 

\paragraph{\texttt{InteractionModule}:}
is the core local interaction backbone used in the reference GEqTrain stack. It operates on a directed radius graph and maintains, for each edge \((i,j)\), a pair of latent states: an invariant edge state \(h_{ij}^{(\ell)}\in\mathbb R^{d_h}\) and an equivariant edge state \(\mathbf z_{ij}^{(\ell)}\), where \(\ell\) denotes the interaction depth.
It constructs an initial mixed scalar--equivariant latent representation from typed attributes and geometry, refines it through a sequence of local interaction layers, and projects the final latent state to the requested output irreps.
The module validates which invariant and equivariant node and edge fields are actually present in \texttt{irreps\_in}, computes the dimension of any optional conditioning tensors, and precomputes the irreps required by the tensor-product path of each layer.

\paragraph{Initial latent construction.}
The initial invariant edge latent is built by concatenating the radial embedding with any available invariant edge attributes and the invariant attributes of the source and target nodes,
\[
\widetilde{h}_{ij}^{(0)}
=
\rho_{ij}
\;\|\;
a_{ij}
\;\|\;
a_i
\;\|\;
a_j,
\]
where absent terms are simply omitted. The initial equivariant edge latent is constructed analogously from the spherical-harmonic edge embedding together with any available equivariant edge attributes and source/target node equivariant attributes,
\[
\widetilde{\mathbf z}_{ij}^{(0)}
=
\mathbf y_{ij}^{\mathrm{sh}}
\;\|\;
a_{ij,\mathrm{eq}}
\;\|\;
a_{i,\mathrm{eq}}
\;\|\;
a_{j,\mathrm{eq}}.
\]
Because concatenation of irreps-carrying tensors may require a canonical ordering, the implementation applies a precomputed permutation before projection. The pair \((\widetilde{h}_{ij}^{(0)},\widetilde{\mathbf z}_{ij}^{(0)})\) is then mapped to the recurrent latent state \((h_{ij}^{(0)},\mathbf z_{ij}^{(0)})\) by an \texttt{EquivariantScalarMLP}, using \texttt{SO3\_Linear} for the equivariant projection and returning the equivariant output in channel-wise format.

\paragraph{Conditioning and output irreps.}
\texttt{InteractionModule} supports conditioning on node or edge fields; it computes the corresponding dimension by concatenating edge-level conditioning directly and node-level conditioning twice, once for the source node and once for the target node.

\paragraph{\texttt{InteractionLayer}:}
Each \texttt{InteractionLayer} takes the current edge states \((h_{ij}^{(\ell)},\mathbf z_{ij}^{(\ell)})\) and produces updated states \((h_{ij}^{(\ell+1)},\mathbf z_{ij}^{(\ell+1)})\). The first step is to build an equivariant edge contribution to the local environment by applying an \texttt{EquivariantScalarMLP} to the current latent pair,
\[
\mathbf e_{ij}^{(\ell)}
=
\Psi_{\mathrm{env}}\!\left(h_{ij}^{(\ell)},\mathbf z_{ij}^{(\ell)}; c_{ij}\right),
\]
where \(c_{ij}\) denotes the optional conditioning tensor. These edge contributions are then aggregated over all outgoing edges of the same source node using \texttt{scatter\_sum},
\[
\mathbf u_i^{(\ell)}
=
\sum_{j\in\mathcal N(i)} \mathbf e_{ij}^{(\ell)}.
\]
Optionally, the aggregated node environment is optionally refined through a MACE-style \cite{Batatia2022MACE} equivariant product block in the intermediate layers, and is then normalized by \texttt{SO3\_LayerNorm}. Finally, the normalized node environment is broadcast back to each outgoing edge through its source node, yielding an edge-local environment \(\mathbf u_{ij}^{(\ell)}=\mathbf u_i^{(\ell)}\).

\paragraph{Optional attention weighting.}
When attention is enabled, the edge contributions \(\mathbf e_{ij}^{(\ell)}\) are reweighted before node aggregation. Queries are generated from the invariant attributes of the source node and keys from the current scalar edge state, both reshaped into \((m,d)\) head structure, where \(m\) is the equivariant multiplicity and \(d\) is the attention head dimension. The attention logits are
\[
\omega_{ijm}^{(\ell)}
=
\frac{\langle Q_{ijm}^{(\ell)},K_{ijm}^{(\ell)}\rangle}{\sqrt{d}},
\]
optionally clipped to a symmetric interval, and normalized with a softmax over edges sharing the same source node. The resulting weights multiply the edge environment contributions before summation. Thus, attention does not replace local aggregation, but modulates the contribution of each edge to its source-centered environment.

\paragraph{Tensor-product interaction.}
The interaction step combines the current equivariant edge state with the broadcast source-centered environment through a tensor product,
\[
\mathbf t_{ij}^{(\ell)}
=
\mathcal T\!\left(\mathbf z_{ij}^{(\ell)}, \mathbf u_{ij}^{(\ell)}\right).
\]
In the implementation this tensor product is realized by an Allegro-style \texttt{Contracter} with precomputed instructions restricted to those output irreps for which a valid path exists. The resulting tensor-product output is subsequently normalized by a second \texttt{SO3\_LayerNorm}. A final \texttt{EquivariantScalarMLP} then maps the normalized tensor-product features to a new invariant state and a new equivariant state,
\[
(\widehat{h}_{ij}^{(\ell+1)},\widehat{\mathbf z}_{ij}^{(\ell+1)})
=
\Psi_{\mathrm{proj}}\!\left(\mathbf t_{ij}^{(\ell)}; c_{ij}\right).
\]
For intermediate layers this projection returns a channel-wise equivariant latent of the same recurrent type, whereas the last interaction layer projects to the pruned final equivariant latent required by the output head.

\paragraph{Residual latent updates.}
Invariant latents are updated through a variance-preserving residual stream. If \(\gamma_\ell\) denotes the learnable residual update coefficient for layer \(\ell\), the implementation combines old and new latents as
\[
h_{ij}^{(\ell+1)}
=
\frac{1}{\sqrt{1+\gamma_\ell^2}}\, h_{ij}^{(\ell)}
+
\frac{\gamma_\ell}{\sqrt{1+\gamma_\ell^2}}\, \widehat{h}_{ij}^{(\ell+1)}.
\]
The coefficient \(\gamma_\ell\) is obtained by applying a sigmoid to a learnable parameter and scaling it by \texttt{residual\_update\_max}. No residual update is applied in the first layer, because the initial latent and first updated latent need not share the same dimensionality. Equivariant residual updates are applied only when explicitly enabled and only when the new equivariant state has the same shape as the previous one. 

\paragraph{Final output projection.}
After the recurrent stack has produced the final latent state \((h_{ij}^{(L)},\mathbf z_{ij}^{(L)})\), a last \texttt{EquivariantScalarMLP} maps it to the requested output field in flat form,
\[
o_{ij}
=
\Psi_{\mathrm{out}}\!\left(h_{ij}^{(L)},\mathbf z_{ij}^{(L)}; c_{ij}\right),
\]
and stores the result under \texttt{out\_field}.

\paragraph{Readout modes.}
\texttt{ReadoutModule} supports both single-field and split invariant/equivariant inputs and outputs. For scalar prediction, the invariant part of the latent representation is processed by an MLP,
\[
\hat{\mathbf{y}}^{(0)} = \mathrm{MLP}(\mathbf{s}),
\]
while equivariant channels, when requested, are mapped through irrep-preserving linear operators,
\[
\hat{\mathbf{y}}^{(l>0)} = \mathrm{EqLin}(\mathbf{u}).
\]
A useful implementation detail is that the module can write both a full mixed output field and a scalar-only auxiliary field. In the scalar-only variant used for the present paper, this flexibility is not required and the final readout is simply \texttt{out\_irreps: 1x0e}.

\paragraph{Chunked execution for large local graphs.}
The strict locality of node- and edge-level stacks also permits inference to be decomposed into memory-bounded subgraphs, with each chunk evaluating a subset of center nodes and their complete local receptive fields. The implementation and empirical memory--runtime scaling of this strategy are described in Section~\ref{sec:chunked_execution}.

\subsection*{Readout and Graph-Level Heads}

\paragraph{\texttt{ReadoutModule}:}
maps node/edge/graph features to target fields, supporting scalar and equivariant outputs, optional conditioning, optional residual mixing, and scalar bias.
For scalar channel with conditioning vector $\mathbf{c}$:
\[
\mathbf{s}'=\gamma(\mathbf{c})\odot\mathbf{s}+\beta(\mathbf{c}),
\qquad
\hat{\mathbf{y}}^{(0)}=\mathrm{MLP}(\mathbf{s}').
\]
Equivariant outputs are produced by equivariant linear maps with either internal weights or weights generated from scalar/conditioning features:
\[
\hat{\mathbf{y}}^{(l>0)} = \mathrm{EqLin}\!\left(\mathbf{u};\,W(\mathbf{s}'\;\text{or}\;\mathbf{c})\right).
\]

\paragraph{\texttt{NodewiseReduce}:}
for graph-level prediction from node outputs:
\[
\mathbf{g}_b = \sum_{i:\,\mathrm{batch}(i)=b} \mathbf{x}_i,
\]
with optional concatenation of an auxiliary residual field before pooling and optional learnable bias.


\section{Implementation Details for Generative Modules}
\label{sec:supp_generative_modules}

\paragraph{Discrete diffusion module.}
In addition to the flow-matching front-end described in the main text, GEqDiff also provides a discrete-time diffusion module through \texttt{ForwardDiffusionModule}. This module samples an integer timestep \(t \in \{0,\dots,T_{\mathrm{train}}-1\}\), encodes it through a sinusoidal embedding, and applies the corresponding noising coefficients \(\alpha_t\) and \(\sigma_t\) returned by a configurable \texttt{NoiseScheduler}. In the current implementation, atomic coordinates are centered graph-wise before corruption, Gaussian noise is sampled and centered in the same way, and the perturbed coordinates are constructed as
\[
\mathbf{x}_t = \alpha_t \mathbf{x} + \sigma_t \boldsymbol{\epsilon},
\]
with the training target stored as the sampled noise \(\boldsymbol{\epsilon}\). The module also registers scalar conditioning fields such as the timestep embedding and diffusion coefficients.

\paragraph{Sampling schedulers for diffusion and flow matching.}
GEqDiff provides scheduler modules for both denoising diffusion and flow-matching objectives. These modules sample a time variable, compute the corresponding data and noise scales, corrupt the selected input fields, and write the associated training targets back into the data dictionary. In the flow-matching case, the target is the scheduler velocity, i.e. the derivative of the interpolated state with respect to the sampled time variable.

For a field \(x^{(f)}\), the scheduler defines an interpolated corrupted state
\[
x^{(f)}_{\tau}
=
\alpha(\tau)\,x^{(f)} + \sigma(\tau)\,\epsilon^{(f)},
\]
where \(\epsilon^{(f)}\) is field-specific noise and \(\alpha(\tau)\), \(\sigma(\tau)\) are the data and noise scales returned by the scheduler. The corresponding flow-matching supervision target is
\[
u^{(f)}_{\tau}
=
\dot{\alpha}(\tau)\,x^{(f)}
+
\dot{\sigma}(\tau)\,\epsilon^{(f)}.
\]
When a binary corruption mask \(M^{(f)}\) is provided, these quantities are applied only to the selected entries:
\[
\tilde{x}^{(f)}_{\tau}
=
M^{(f)} \odot x^{(f)}_{\tau}
+
\left(1-M^{(f)}\right)\odot x^{(f)},
\qquad
\tilde{u}^{(f)}_{\tau}
=
M^{(f)} \odot u^{(f)}_{\tau}.
\]
Thus, unmasked entries remain fixed by default and receive a zero target. If no mask is specified, \(M^{(f)}=1\) and the whole field is corrupted. Optional centering can be applied before corruption, and noise can be centered independently, allowing the reference frame and the noised region to be controlled separately.

The field-wise noising/interpolation process is configured through the \texttt{corrupt\_fields} list, as shown in Listing~\ref{lst:supp_lego_flow_corruption}. Each entry specifies the input field to perturb, the output velocity target to generate, and optional masking or centering rules. Different fields can therefore be transported with different preprocessing choices within the same forward module. In the LEGO benchmark, coordinates, shape descriptors, and dipole-like vectors are perturbed jointly but with separate target fields. Coordinates are centered using \texttt{pocket\_mask} to define the fixed context, while coordinate noise is centered over the generated region specified by \texttt{ligand\_mask}. The shape and dipole fields are instead perturbed only on generated nodes without additional centering. This makes the conditional generation task explicit: the model observes the fixed scaffold/pocket fields and learns to predict the flow velocities for the masked generated/ligand fields.

\begin{listing}[h]
\centering
\caption{Field-wise noising/interpolation configuration used for the LEGO flow-matching task.}
\label{lst:supp_lego_flow_corruption}
\begin{lstlisting}[language=yaml,gobble=4]
    - _target_: geqdiff.nn.ForwardFlowMatchingModule
      name: flow_matching
      Tmax: 100
      flow_target_parameterization: scheduler_velocity
      flow_time_parameterization: tau
      t_embedder_kwargs:
        embedding_dim: 64
      corrupt_fields:
        - field: pos
          out_field: velocity
          center: true
          center_mask_field: pocket_mask
          center_noise: true
          noise_center_mask_field: ligand_mask
          mask_field: ligand_mask
    
        - field: shape_features
          out_field: shape_features_velocity
          center: false
          mask_field: ligand_mask
    
        - field: dipole_direction
          out_field: dipole_direction_velocity
          center: false
          mask_field: ligand_mask
\end{lstlisting}
\end{listing}

The output side mirrors the same field-wise structure. Each transported field is assigned a readout head whose output irreps match the representation of the corresponding velocity target. Coordinates and dipole directions use vector-valued \texttt{1x1o} heads, while the shape descriptor uses the mixed representation transported in the benchmark.

\begin{listing}[h]
\centering
\caption{Readout heads used to predict field-specific flow velocities in the LEGO flow-matching task.}
\label{lst:supp_lego_flow_readouts}
\begin{lstlisting}[language=yaml,gobble=4]
    - _target_: geqtrain.nn.ReadoutModule
      name: position_head
      field: node_features
      out_field: velocity
      out_irreps: 1x1o

    - _target_: geqtrain.nn.ReadoutModule
      name: shape_head
      field: node_features
      out_field: shape_features_velocity
      out_irreps: 1x0e + 1x1o + 1x2e + 1x3o

    - _target_: geqtrain.nn.ReadoutModule
      name: dipole_direction_head
      field: node_features
      out_field: dipole_direction_velocity
      out_irreps: 1x1o
\end{lstlisting}
\end{listing}


\section{Tensorial Shielding Prediction on the ShiftML3 Dataset}
\label{si:shiftml3_tensor_protocol}

\subsection{Dataset and evaluation split}
\label{si:shiftml3_dataset}

To test tensorial target prediction in GEqTrain, we used the tensor-containing molecular-solid dataset introduced with ShiftML3~\cite{Kellner2025}. The dataset contains DFT-computed chemical shielding tensors for molecular crystals sampled from the Cambridge Structural Database, together with thermally distorted structures. In the present supplementary analysis, we report results for the H, C, N, and O nuclei used in the main chapter comparison.

We evaluate the model on the ShiftML3 \texttt{CSD-test} split. This is the hold-out test set defined by the ShiftML3 authors, constructed to avoid leakage between relaxed and thermally distorted structures derived from the same CSD identifier.

\subsection{GEqTrain configuration for NMR experiments}
\label{si:shiftml3_geqtrain_config}

The model was trained with a joint scalar--tensor objective. The scalar head predicts the isotropic shielding target, while the tensorial head predicts the non-scalar irreducible shielding components. The loss is
\begin{equation}
    \mathcal{L}
    =
    \mathcal{L}_{\mathrm{iso}}
    +
    10\,\mathcal{L}_{\mathrm{tensor}},
\end{equation}
where both terms use a LogCosh loss. Metrics are reported as MAE and RMSE per atomic species and are computed only on the masked center atoms used as prediction targets.

The model uses a local cutoff of \(6.0\,\text{\AA}\), 16 radial basis functions, and spherical harmonics up to \(l_{\max}=2\). Scalar node and edge embeddings use 512 channels. Equivariant edge features include \(8 \times l=1\) and \(8 \times l=2\) channels. The interaction block uses
two message-passing layers, attention-based aggregation, a latent dimension of 512, equivariant latent multiplicity 8, and outputs scalar, vector, and rank-2 irreducible features. Two readout heads are used:
\begin{align}
    \texttt{cs\_iso}    &: 1 \times 0e, \\
    \texttt{cs\_tensor} &: 1 \times 1e + 1 \times 2e .
\end{align}
Training used AdamW with learning rate \(10^{-4}\), exponential learning-rate decay with \(\gamma=0.98\), batch size 2, validation batch size 4, and a maximum of 250 epochs.

\subsection{Irreducible tensor-component errors}
\label{si:shiftml3_tensor_results}

Table~\ref{tab:shiftml3_tensor_irrep} reports RMSEs for the non-scalar shielding components on the ShiftML3 \texttt{CSD-test} split. To make the connection with the ShiftML3 tensor terminology explicit, we label the \(l=1\) block as the antisymmetric sector and the \(l=2\) block as the symmetric traceless sector. The values, however, are computed directly in the irreducible representation used for training.

For a nucleus type \(Z\) and irreducible order \(l\), the reported error is
\begin{equation}
    \mathrm{RMSE}_{Z,l}
    =
    \sqrt{
    \frac{1}{N_Z(2l+1)}
    \sum_{i:z_i=Z}
    \sum_{m=1}^{2l+1}
    \left(
        \hat{T}^{(l)}_{i,m} - T^{(l)}_{i,m}
    \right)^2
    },
\end{equation}
where \(T^{(l)}_{i,m}\) denotes the target irreducible shielding component of atom \(i\), after applying the same inverse normalization used for evaluation of physical-unit errors. All values are in ppm.

\begin{table}[htbp]
    \centering
    \caption{RMSEs for non-scalar irreducible shielding components on the ShiftML3 \texttt{CSD-test} split. The labels connect the GEqTrain irreducible blocks to the corresponding tensor sectors: the \(l=1\) block represents the antisymmetric sector, while the \(l=2\) block represents the symmetric traceless sector. Errors are computed in the native irreducible representation used for training and are reported in ppm.}
    \label{tab:shiftml3_tensor_irrep}
    \begin{tabular}{lcc}
        \toprule
        \textbf{Nucleus} &
        \textbf{RMSE \(\sigma_{\mathrm{antisym}}\) / \(l=1\)} &
        \textbf{RMSE \(\sigma_{\mathrm{sym,tr}}\) / \(l=2\)} \\
        \midrule
        $^{1}$H  & 0.72 & 0.96 \\
        $^{13}$C & 4.38 & 5.00 \\
        $^{15}$N & 9.83 & 15.08 \\
        $^{17}$O & 12.80 & 20.17 \\
        \bottomrule
    \end{tabular}
\end{table}

This table complements the isotropic shielding results in the main text. The scalar benchmark evaluates the usual NMR target \(\sigma_{\mathrm{iso}}\), whereas the tensorial analysis tests whether the same GEqTrain configuration can learn local observables with prescribed non-scalar transformation behavior.


\section{LEGO benchmark}
\label{sec:supp_lego}

This section describes the synthetic LEGO benchmark used to evaluate mixed-field equivariant generation in GEqDiff.
The benchmark is designed as a controlled test case to assess whether a single typed equivariant architecture can jointly generate Cartesian positions, higher-order shape descriptors, and vector-valued attributes.

Each generated assembly is represented as a graph
\[
\mathcal{G} = (\mathcal{V}, \mathcal{E}),
\]
where each node \(i \in \mathcal{V}\) corresponds to a LEGO-like brick or brick element.
Each node carries three target fields:
\[
x_i = \left(\mathbf{r}_i, \mathbf{s}_i, \boldsymbol{\mu}_i\right),
\]
where \(\mathbf{r}_i \in \mathbb{R}^3\) is the anchor position,
\(\mathbf{s}_i \in \mathbb{R}^{16}\) is a shape descriptor, and
\(\boldsymbol{\mu}_i \in \mathbb{R}^3\) is a dipole-like vector.
The shape descriptor is stored as a 16-dimensional coefficient vector corresponding to the representation content
\[
\texttt{1x0e + 1x1o + 1x2e + 1x3o}.
\]
The dipole-like vector is treated as a direct vector-valued target, with both orientation and magnitude encoded in \(\boldsymbol{\mu}_i\).

\subsection{Discrete procedural generation}
\label{sec:supp_lego_generation_principle}

LEGO structures are generated on a discrete three-dimensional lattice.
A turtle-like construction rule sequentially places brick anchors and updates a local orthonormal frame using axis-aligned rotations from the cubic grid symmetry group.
This makes connectivity and relative orientation explicit, while keeping the generation process simple and reproducible.

Extended bricks such as \texttt{1x2}, \texttt{L}-shaped, and \texttt{T}-shaped elements are represented by finite voxel footprints.
During generation, each proposed placement is checked against a global occupancy grid.
Placements that would introduce overlaps are rejected, and failed trajectories are resampled.
Thus, all accepted structures are connected and overlap-free by construction.

\subsection{Procedural scaffold grammars}
\label{sec:supp_lego_scaffolds}

The structures are sampled from three simple grammars inspired by secondary-structure-like motifs:
\begin{itemize}
    \item \textbf{Beta-sheet-like scaffolds:} Compact, locally planar assemblies formed by alternating runs of \texttt{1x1} and \texttt{1x2} bricks, punctuated by tight \(180^\circ\) U-turns implemented using paired \texttt{L-shape} bricks. The parity of the turns is strictly alternated to fold the beta-sheet densely without self-intersection.
    \item \textbf{Alpha-helix-like scaffolds:} Discrete helical assemblies built from a deterministic periodic brick grammar. The helix generator uses a fixed local phase progression and chiral frame updates so that each sample follows a consistent alpha-helix-like pattern.
    \item \textbf{Mixed scaffolds:} Heterogeneous assemblies created by alternately sequencing beta-sheet-like and alpha-helix-like segments. Because the discrete turtle maintains a strict absolute frame, the transition between domains is contiguous, with the first element of the new motif initialized from the terminal position and local frame of the preceding segment.
\end{itemize}

The precise sampling parameters used for the final experiments are reported in Table~\ref{tab:supp_lego_generation_params}.
These parameters control the distribution over scaffold classes and the number of nodes.

\begin{table}[t]
\centering
\caption{
Generation parameters for the deterministic scaffold LEGO benchmark.
}
\label{tab:supp_lego_generation_params}
\begin{tabular}{lc}
\toprule
Parameter & Value/range \\
\midrule
Scaffold classes & beta-sheet-like, alpha-helix-like, mixed \\
Number of source assemblies (train, valid, test) & 1,000, 50, 100 \\
Number of extracted conditional examples per source & 5 \\
Node-count range & 8-40 \\
Diffused-subassembly size range & 4-12 \\
\bottomrule
\end{tabular}
\end{table}

\subsection{Roles, shape descriptors, and dipoles}
\label{sec:supp_lego_features}

The exact structural role \(\rho_i\) and local affine frame of every node are intrinsically determined by the L-system's generative path. These roles denote specific local motifs, such as terminal elements, beta-sheet bodies, helical bodies, or turn junctions.

For each node \(i\), the known role and frame deterministically dictate a target discrete brick type \(b_i\) and orientation \(q_i\). Rather than using these discrete labels directly as generative targets, they are mapped to a continuous shape descriptor \(\mathbf{s}_i \in \mathbb{R}^{16}\). This descriptor corresponds to the spherical-harmonic signature of the brick's exposed faces, yielding a continuous, higher-order equivariant representation:
\[
(b_i, q_i, \mathbf{s}_i) = f_{\mathrm{shape}}(\rho_i, \mathbf{t}_i, \mathbf{n}_i, \mathbf{b}_i),
\]
where \((\mathbf{t}_i, \mathbf{n}_i, \mathbf{b}_i)\) is the local orthonormal frame derived from the sequence tangent, normal, and binormal.

Similarly, a dipole-like vector \(\boldsymbol{\mu}_i \in \mathbb{R}^3\) is directly calculated based on the role and local frame:
\[
\boldsymbol{\mu}_i = f_{\mathrm{dipole}}(\rho_i, \mathbf{t}_i, \mathbf{n}_i, \mathbf{b}_i).
\]
This controlled directional attribute tests whether the generative model can preserve vector-valued information coupled to local geometry. This formulation ensures that the target signals for both the 16-dimensional shape channel and the 3D vector-valued dipole are mathematically consistent with the underlying discrete assembly.

\subsection{Conditional reconstruction and flow matching}
\label{sec:supp_lego_task}

The learning task is formulated as conditional reconstruction of a connected subassembly.
For each complete assembly, a connected subset of nodes \(\mathcal{L} \subset \mathcal{V}\) is selected as the diffused region, while the complement \(\mathcal{C} = \mathcal{V} \setminus \mathcal{L}\) acts as fixed context. For each diffused node \(i \in \mathcal{L}\), the model must jointly reconstruct its position, shape, and dipole: \( (\mathbf{r}_i, \mathbf{s}_i, \boldsymbol{\mu}_i) \).

The benchmark uses direct flow matching over these three mixed-field channels. For each target field \(y \in \{\mathbf{r}, \mathbf{s}, \boldsymbol{\mu}\}\), we define a linear interpolation between a noise sample \(y_0\) and the data sample \(y_1\):
\[
y_t = (1-t)y_0 + t y_1,
\qquad t \in [0,1],
\]
with target velocity \(u_t = y_1 - y_0\). The training objective is the sum of channel-wise velocity-matching losses:
\[
\mathcal{L}_{\mathrm{LEGO}}
=
\lambda_{\mathbf{r}}
\mathcal{L}_{\mathbf{r}}
+
\lambda_{\mathbf{s}}
\mathcal{L}_{\mathbf{s}}
+
\lambda_{\boldsymbol{\mu}}
\mathcal{L}_{\boldsymbol{\mu}},
\]
where each term minimizes the squared \(L_2\) error between the network's predicted velocity \(v_\theta(y_t, t, \mathcal{G}, \mathcal{C})\) and the target \(u_t\).

\subsection{Decoding and visualization}
\label{sec:supp_lego_sampling_decoding}

At sampling time, the flow-matching velocity field is integrated from \(t=0\) to \(t=1\) to obtain continuous generated predictions: \( (\hat{\mathbf{r}}_i, \hat{\mathbf{s}}_i, \hat{\boldsymbol{\mu}}_i) \). Because the shape channel \(\hat{\mathbf{s}}_i\) is an equivariant tensor of spherical-harmonic coefficients, we employ two complementary methods to interpret and visualize it:

\begin{itemize}
    \item \textbf{Continuous Surface Rendering:} The predicted spherical-harmonic coefficients \(\hat{\mathbf{s}}_i\) directly parameterize a smooth radial surface field around each node. Rendering these surfaces allows for direct visual inspection of the flow-matching trajectory, showing how the continuous shape representation smoothly evolves across steps without discrete approximation artifacts.
    \item \textbf{Discrete Brick Decoding:} For quantitative geometric evaluation and final macroscopic visualization, the continuous shape tensor is mapped back to a rigid building block. This is achieved via nearest-neighbor matching in the shape-feature space against a library of prototype exemplars \(\mathcal{P}\):
    \[
    (\hat{b}_i, \hat{q}_i)
    =
    \operatorname*{argmin}_{(b,q) \in \mathcal{P}}
    \left\|
    \hat{\mathbf{s}}_i - \mathbf{s}(b,q)
    \right\|_2.
    \]
    This decoding step yields a discrete brick type \(\hat{b}_i\) and orientation \(\hat{q}_i\), which are plotted as exact interlocking meshes to assess the structural validity and physical clashes of the final assembly.
\end{itemize}

This dual representation cleanly separates the continuous geometric tensors learned by the network from the discrete macroscopic structures they encode.

\subsection{Evaluation metrics}
\label{sec:supp_lego_metrics}

All LEGO samples are evaluated in two anchor modes.  In the \emph{raw} mode,
metrics are computed from the continuous anchors produced by the sampler.  In the
\emph{voxelized} mode, sampled anchors are first snapped to the nearest lattice
site, while all other decoded quantities are kept unchanged.  The voxelized mode
is the default reporting mode because the task is defined on a discrete LEGO
lattice and small continuous coordinate errors should not be confused with
topological errors.

Let $\mathcal{L}$ denote the masked/designable bricks and let the unmasked bricks
be the fixed context.  For paired evaluations, the generated structure is scored
against its corresponding reference structure.  The score card contains absolute
validity, relative validity, shape fidelity, dipole fidelity, and pose fidelity.

\paragraph{Absolute validity.}
Validity measures intrinsic geometric soundness of the decoded assembly, without
forcing it to match the reference.  Bricks are expanded into their occupied unit
voxels after applying the decoded type and rotation.  For each pair of occupied
voxels with displacement $\Delta$, the continuous overlap contribution is
\[
v(\Delta)
=
\prod_{a \in \{x,y,z\}}
\max(0, 1 - |\Delta_a|).
\]
Pair overlaps are summed over all voxel pairs belonging to two different bricks.
To avoid over-penalizing numerical near misses, each brick-pair overlap is reduced by a tolerance $\tau_{\mathrm{ov}}=0.01$ before contributing to the effective overlap,
\[
V_{\mathrm{eff}}
=
\sum_{i<j} \max(0, V_{ij}-\tau_{\mathrm{ov}}).
\]
Pairs with $V_{ij}>0.08$ are counted as severe overlaps.  Connectivity is computed from face-contact edges whose continuous contact area is larger than $0.12$.
If $K$ is the number of connected components and $N_{\mathrm{sev}}$ is the number of severe-overlap pairs, the implemented validity score is
\[
S_{\mathrm{valid}}
=
100
\frac{
\exp(-12.0\,V_{\mathrm{eff}} - 0.9\,N_{\mathrm{sev}})
}{
\max(1,\sqrt{K})
}.
\]
A structure is additionally flagged as ``valid-like'' when
$V_{\mathrm{eff}}\le 0.02$, $N_{\mathrm{sev}}=0$, and $K=1$.

\paragraph{Dipole fidelity.}
Dipoles are represented as direct three-dimensional vectors, with the vector
direction encoding orientation and polarity and the norm encoding interaction
strength. The cosine diagnostic is averaged over the designable set
after normalizing nonzero vectors. If both sampled and reference dipoles are
zero, the pair contributes a cosine of $1$; if exactly one is zero, it contributes $0$. The magnitude diagnostic is
\[
\mathrm{RMSE}_{|\mu|}
=
\left[
\frac{1}{|\mathcal{L}|}
\sum_{i\in\mathcal{L}}
\left(\|\hat{\boldsymbol{\mu}}_i\|_2-\|\boldsymbol{\mu}_i\|_2\right)^2
\right]^{1/2}.
\]
The scalar score uses three terms: direction agreement $C_\mu=(1+\overline{\cos}_{\mu})/2$, magnitude agreement $M_\mu=\exp[-(\mathrm{RMSE}_{|\mu|}/0.28)^2]$, and an energy-alignment term.
The energy term compares the sampled and reference contact energies computed
from the same polar-contact energy used during dataset generation.  In the
implementation, the energy field is read as \texttt{weighted\_dipole\_energy}
when present, otherwise as \texttt{dipole\_total\_energy}. With $\Delta E_\mu = E_\mu^{\mathrm{sample}}-E_\mu^{\mathrm{ref}}$, the score is
\[
S_\mu
=
100\left[
0.60 C_\mu
+0.20 M_\mu
+0.20 \exp(-|\Delta E_\mu|/1.5)
\right].
\]
The reported diagnostics include the mean cosine, angular error in degrees,
magnitude RMSE, energy difference, and energy-alignment term.

\paragraph{Pose fidelity.}
Pose fidelity measures whether designable anchors return to the paired reference
positions while fixed-context anchors remain unchanged. Let $d_{\mathcal{L}}^{\mathrm{mean}}$ and $d_{\mathcal{L}}^{\max}$ be the mean and
maximum displacement of designable anchors, and let $d_{\mathrm{fixed}}^{\max}$
be the maximum displacement of fixed-context anchors. The implemented score is
\[
S_{\mathrm{pose}}
=
100
\exp\left(
-d_{\mathcal{L}}^{\mathrm{mean}}/0.50
-d_{\mathcal{L}}^{\max}/1.20
-1.2\,d_{\mathrm{fixed}}^{\max}
\right).
\]

\section{Computational Scaling and Execution Strategies}
\label{sec:computational_scaling}

GEqTrain provides two complementary mechanisms for computational scaling.
Distributed Data Parallel training increases throughput by replicating the model across GPUs and distributing minibatches, whereas chunked execution reduces the peak memory required to evaluate a single large local graph.
These mechanisms therefore address training-throughput scaling and single-system memory scaling, respectively.

\subsection{Distributed Multi-GPU Training with PyTorch DDP}
\label{sec:ddp}

GEqTrain uses PyTorch Distributed Data Parallel (DDP) for synchronous data-parallel training across multiple GPUs. Each process owns a complete model replica and processes a distinct portion of the minibatch, with gradients synchronized before each optimization step.
DDP can reduce wall-clock training time and increase the effective batch size..
This distributed capability is activated by passing the \texttt{--ddp} flag to the training script. The framework supports two primary methods for launching distributed runs: using \texttt{torchrun} for single multi-GPU machines and using SLURM for High-Performance Computing (HPC) clusters.

\paragraph{Using \texttt{torchrun} on a Single Machine}
For a single workstation or server with multiple GPUs, \texttt{torchrun} is the standard and recommended launcher. It automatically manages the setup of the distributed environment.

To start a training run, you specify the number of GPUs to use with the \texttt{--nproc\_per\_node} argument. For example, to train on 4 GPUs, the command is:
\begin{lstlisting}[language=bash,numbers=none]
torchrun --nproc_per_node=4 \
    geqtrain/scripts/train.py \
    path/to/conf.yaml \
    --ddp
\end{lstlisting}

\subparagraph{Important Considerations and Tips}
\begin{itemize}
    \item \textbf{Selecting Specific GPUs}: To restrict training to a subset of available GPUs, you can use the \texttt{CUDA\_VISIBLE\_DEVICES} environment variable. For instance, to use only GPUs 0 and 2, you would set \texttt{export CUDA\_VISIBLE\_DEVICES=0,2} before running the \texttt{torchrun} command.
    \item \textbf{NCCL Hangs Workaround}: On some systems, low-level hardware or driver conflicts can cause the training process to hang during initialization. If this occurs, setting the environment variable \texttt{export NCCL\_P2P\_DISABLE=1} forces a more robust, albeit potentially slower, communication path that can resolve the issue.
\end{itemize}

\paragraph{Using SLURM on an HPC Cluster}
On HPC clusters managed by the SLURM workload manager, distributed training is typically launched via a submission script. This approach allows for precise resource allocation and leverages SLURM's infrastructure to manage inter-process communication.

The process involves creating an \texttt{sbatch} script that requests the necessary resources (nodes, GPUs per task, etc.) and sets up the environment. For DDP to work correctly, two environment variables, \texttt{MASTER\_ADDR} and \texttt{MASTER\_PORT}, must be set so that all processes know how to communicate with the master process (rank 0). These can be configured automatically within the SLURM script.

Instead of using \texttt{torchrun}, the training can be launched with \texttt{srun}, explicitly passing the communication details via command-line arguments to the script:
\begin{lstlisting}[language=bash,numbers=none]
# This command would be placed inside an sbatch script
srun geqtrain-train config.yaml \
    --ddp \
    --master-addr $MASTER_ADDR \
    --master-port $MASTER_PORT
\end{lstlisting}
The training script accepts \texttt{--master-addr} and \texttt{--master-port} to facilitate this setup. A complete example of a SLURM submission script (\texttt{train.sbatch}) is provided in the project's \texttt{README.md} file.

\subsection{Memory-Bounded Inference through Chunked Execution}
\label{sec:chunked_execution}

\paragraph{Chunking strategy.}
For strictly local node- and edge-level stacks, GEqTrain can evaluate a single molecular graph as a sequence of memory-bounded subgraphs rather than through one monolithic forward pass. When \texttt{chunking} is enabled, the framework iteratively selects a subset of source, or center, nodes whose induced local neighborhoods fit within the user-defined \texttt{batch\_max\_atoms} limit.
The corresponding subgraph is evaluated, its center-node outputs are retained, and execution proceeds to the remaining centers until the complete graph has been covered.

Because each prediction depends only on the receptive field of its associated center node, this decomposition preserves the semantics of strictly local message passing. Chunking is not applied to graph-level objectives, for which the output or loss can depend jointly on all nodes and cannot in general be reconstructed from independent local evaluations.

\paragraph{Synthetic scaling benchmark.}
We quantified the resulting memory--runtime trade-off using a 28.96-million-parameter GEqTrain interaction model on synthetic sparse graphs with \(N\) nodes and fixed out-degree 32, corresponding to \(32N\) directed edges. Benchmarks were performed on a single NVIDIA RTX A6000 GPU with 47.4~GB of memory; inference times are averages over three executions following one warm-up run.

For monolithic full-graph inference, peak CUDA reserved memory increased approximately linearly with graph size, reaching 44.94~GB at 6,882 nodes and 220,224 edges. The next evaluated size, containing 11,012 nodes, resulted in an out-of-memory error. Once execution was divided into multiple chunks,
peak memory was instead controlled primarily by the chunk budget rather than the total graph size. Both the 1,000- and 2,000-node chunk settings successfully processed graphs containing 28,192 nodes and 902,144 edges, the largest systems tested.

This reduction in peak memory comes at the cost of repeated subgraph construction and serial evaluation. At 6,882 nodes, mean inference time increased from 8.04~s for full-graph execution to 56.39~s and 49.45~s for chunk budgets of 1,000 and 2,000 nodes, respectively. Chunking therefore provides a configurable memory--throughput trade-off and should be interpreted as a mechanism for memory-bounded inference rather than as a runtime optimization. The synthetic fixed-degree graphs isolate scaling with graph size at constant sparsity and are not intended to reproduce the topology or throughput of a particular molecular system.

\begin{figure}[htbp]
    \centering
    \includegraphics[width=0.90\textwidth]{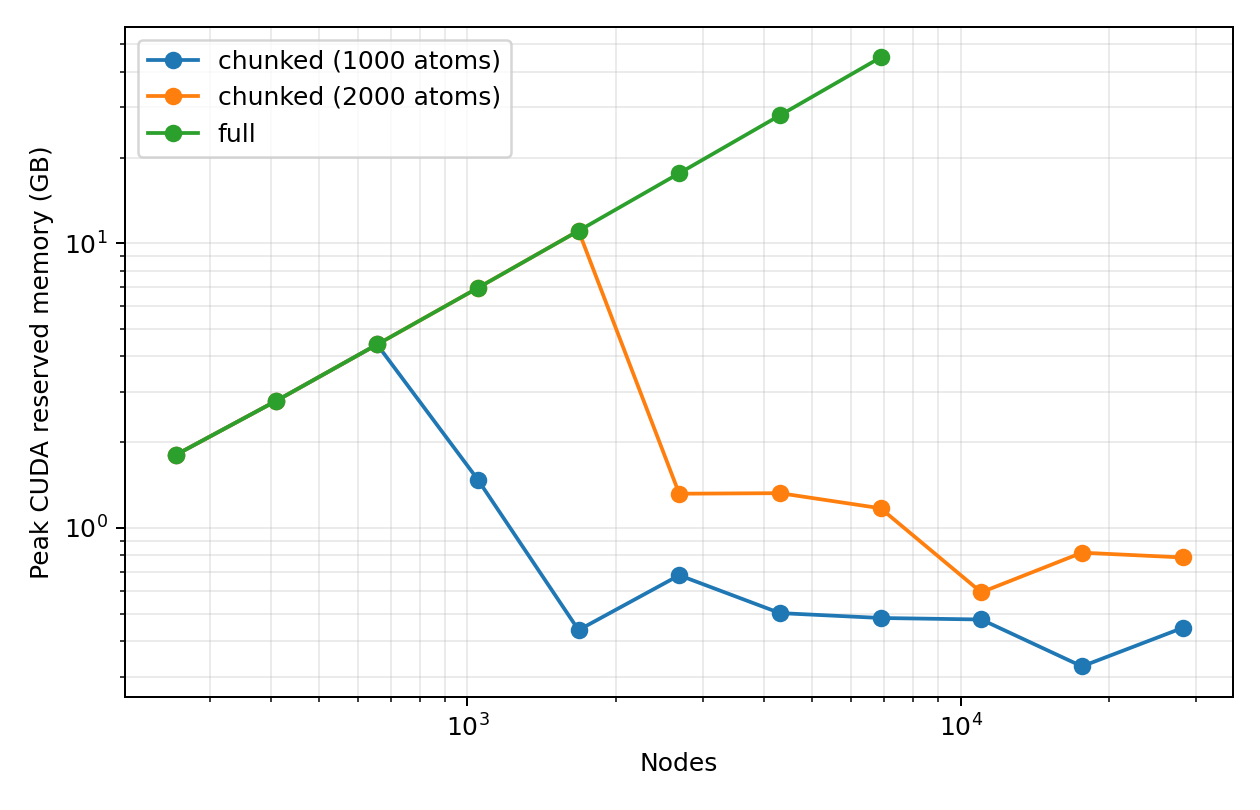}

    \vspace{0.5em}

    \includegraphics[width=0.90\textwidth]{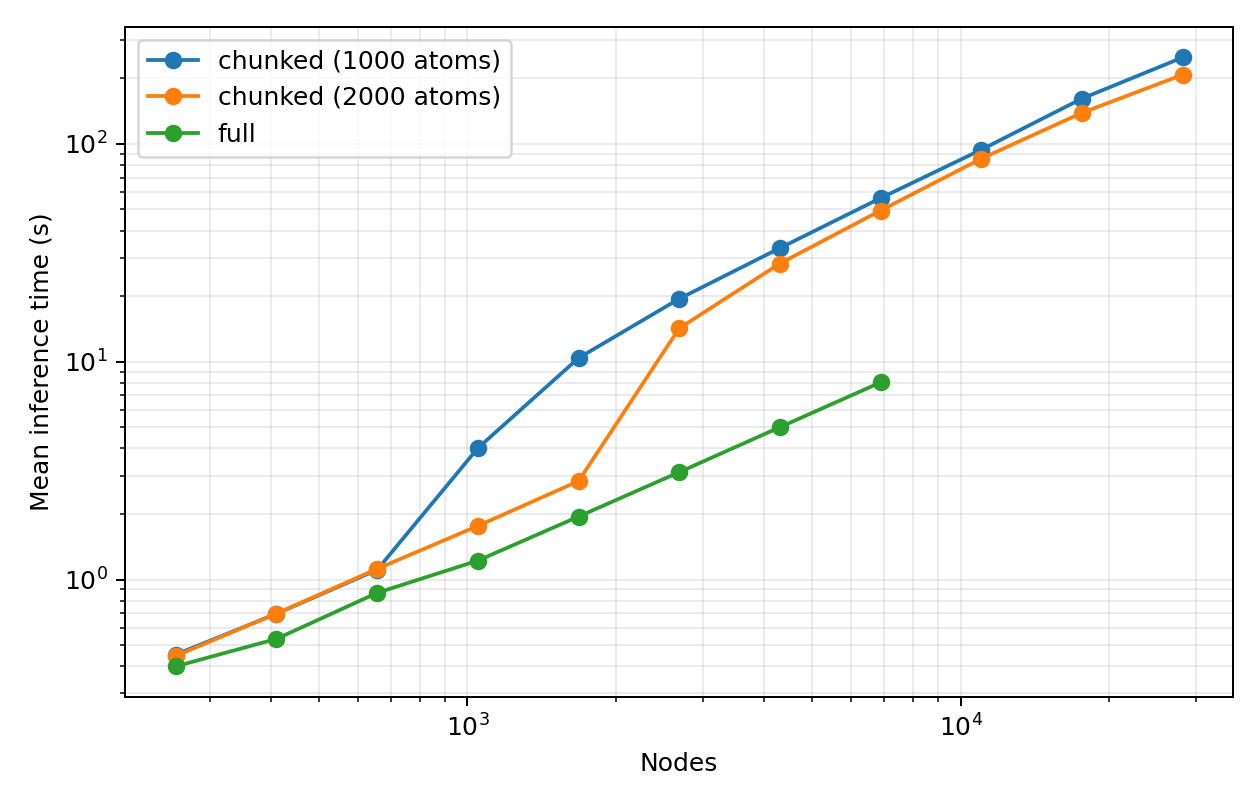}
    \caption{
    \textbf{Memory and runtime scaling of full and chunked GEqTrain inference.}
    A 28.96-million-parameter model was evaluated on synthetic directed graphs with
    fixed out-degree 32 on a single NVIDIA RTX A6000 GPU with 47.4~GB of memory.
    Top: peak CUDA memory reserved by the PyTorch allocator.
    Bottom: mean wall-clock inference time over three executions following one
    warm-up. Before the chunk budget is exceeded, chunked execution reduces to a
    single subgraph and has similar memory requirements to full-graph inference.
    Full-graph memory grows approximately linearly and inference fails with an
    out-of-memory error at 11,012 nodes. Chunked execution bounds peak memory and
    successfully processes 28,192 nodes, the largest graph tested, while introducing
    additional runtime because subgraphs are evaluated sequentially.
    }
    \label{fig:supp_chunk_scaling}
\end{figure}

\printbibliography